\documentclass[sigconf]{acmart}
\usepackage{bm}
\usepackage{xspace} 
\usepackage{algorithm} 
\usepackage{algpseudocode}
\usepackage{multirow}
\usepackage{colortbl}  
\usepackage{eqparbox}
\usepackage{tablefootnote}
\usepackage{balance}

\newcommand{\Figref}[1]{Figure.~\ref{#1}}

\newcommand{\AlgRef}[1]{Algorithm~\ref{#1}}

\newcommand{\Eqref}[1]{Eq.~(\ref{#1})}

\newcommand{\av}{{\boldsymbol a}}
\newcommand{\bv}{{\boldsymbol b}}
\newcommand{\cv}{{\boldsymbol c}}

\newcommand{\ov}{{\boldsymbol o}}

\newcommand{\rv}{{\boldsymbol r}}

\newcommand{\vvec}{{\boldsymbol v}}

\newcommand{\xv}{{\boldsymbol x}}

\newcommand{\Amat}{{\bm A}}

\newcommand{\Rmat}{{\bm R}}

\newcommand{\Vmat}{{\bm V}}
\newcommand{\Wmat}{{\bm W}}

\newcommand{\Transpose}{\mathrm{T}}
\newcommand{\ReLU}{\mathit{ReLU}}
\newcommand{\Sigmoid}{\mathit{Sigmoid}}

\newcommand{\ModelName}{\text{X-GGM}}
\newcommand{\RGenerator}{\text{R-GGM}}
\newcommand{\NGenerator}{\text{N-GGM}}
\newcommand{\VQA}{\mathcal{{VQA}}}
\newcommand{\VQAR}{\mathcal{{VQA_R}}}
\newcommand{\VQAC}{\mathcal{{VQA_C}}}

\copyrightyear{2021}
\acmYear{2021}
\setcopyright{acmcopyright}\acmConference[MM '21]{Proceedings of the 29th ACM International Conference on Multimedia}{October 20--24, 2021}{Virtual Event, China}
\acmBooktitle{Proceedings of the 29th ACM International Conference on Multimedia (MM '21), October 20--24, 2021, Virtual Event, China}
\acmPrice{15.00}
\acmDOI{10.1145/3474085.3475350}
\acmISBN{978-1-4503-8651-7/21/10}

\renewcommand\footnotetextcopyrightpermission[1]{}
\begin{document}
\fancyhead{}

\title{$\ModelName$: Graph Generative Modeling for Out-of-Distribution Generalization in Visual Question Answering}

\author{
   Jingjing Jiang, 
   Ziyi Liu, 
   Yifan Liu, 
   Zhixiong Nan, 
   Nanning Zheng
}
\authornote{Corresponding author.}
\affiliation{%
  \institution{Institute of Artificial Intelligence and Robotics, Xi'an Jiaotong University, Shannxi, China}
}
\email{jingjingjiang2017@gmail.com, {liuziyi,lyf666}@stu.xjtu.edu.cn,{nzx2018,nnzheng}@mail.xjtu.edu.cn}


\begin{abstract}
Encouraging progress has been made towards Visual Question Answering (VQA) in recent years, but it is still challenging to enable VQA models to adaptively generalize to out-of-distribution (OOD) samples. 
Intuitively, recompositions of existing visual concepts (\ie, attributes and objects) can generate unseen compositions in the training set, which will promote VQA models to generalize to OOD samples. 
In this paper, we formulate OOD generalization in VQA as a compositional generalization problem and propose a graph generative modeling-based training scheme (X-GGM) to implicitly model the problem. 
X-GGM leverages graph generative modeling to iteratively generate a relation matrix and node representations for the predefined graph that utilizes attribute-object pairs as nodes. 
Furthermore, to alleviate the unstable training issue in graph generative modeling, we propose a gradient distribution consistency loss to constrain the data distribution with adversarial perturbations and the generated distribution. 
The baseline VQA model (LXMERT) trained with the X-GGM scheme achieves state-of-the-art OOD performance on two standard VQA OOD benchmarks, \ie, VQA-CP v2 and GQA-OOD. 
Extensive ablation studies demonstrate the effectiveness of X-GGM components. 
Code is available at \url{https://github.com/jingjing12110/x-ggm}. 
\end{abstract}


\maketitle

\section{Introduction}


Visual Question Answering (VQA)~\cite{antol2015vqa}, answering questions conditioned on understanding the given image and question, is a task toward general AI~\cite{teney2017visual}. 
Although considerable progress has been made in VQA, most methods usually process similar data distributions between training and test sets.
However, a good VQA model should be robust against answer distribution shift~\cite{kervadec2020roses} and perform well on out-of-distribution (OOD) testing~\cite{teney2020value}, \ie, evaluating the ability of the model to generalize beyond dataset-specific language biases~\cite{agrawal2016analyzing,zhang2016yin,goyal2017making,agrawal2018don}. 
This paper focuses on OOD generalization in VQA, aiming to improve the OOD generalization ability of baseline VQA models while preserving their in-domain (ID) performance.

\begin{figure}[!t]
\begin{center}
\includegraphics[width=1.0\linewidth]{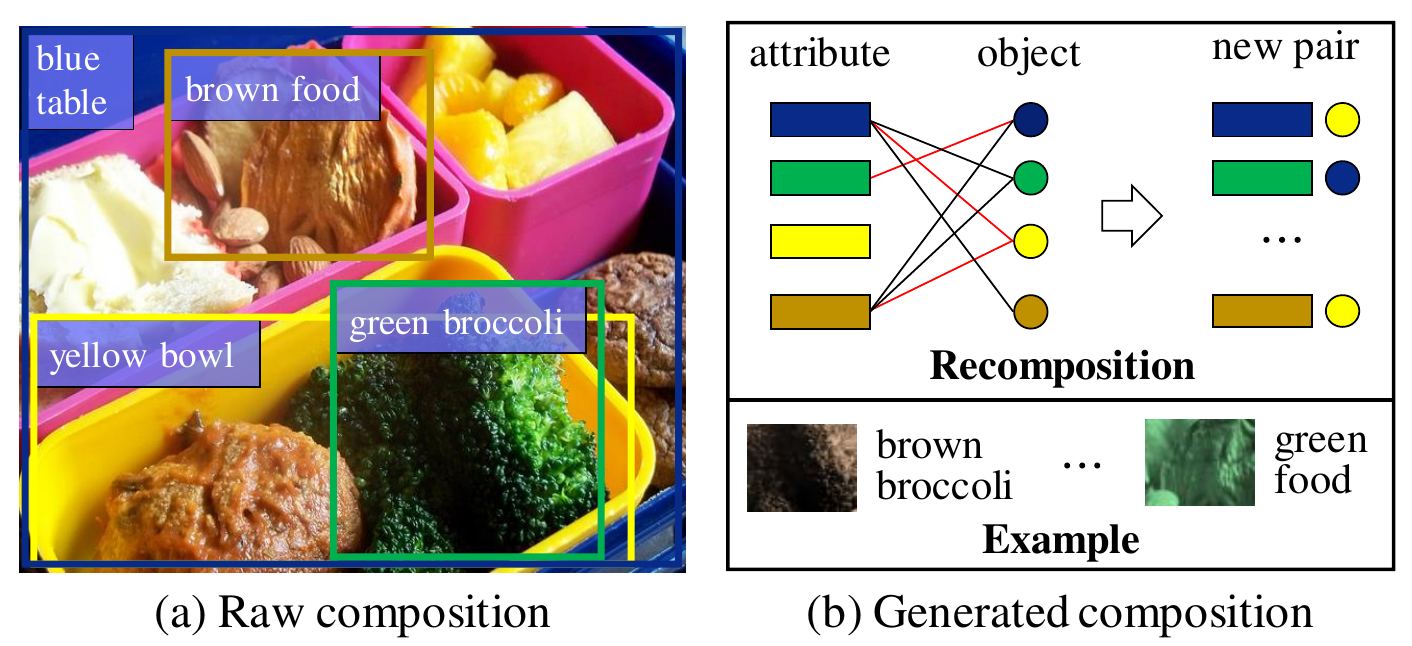}
\end{center}
\vspace{-2mm}
\caption{
Compositional generalization implied in graph generative modeling. 
New compositions (\eg, brown broccoli) of attributes and objects can be generated by raw compositions (\ie, green broccoli and brown food). 
}
\label{fig:motivation} 
\Description{Motivation}
\vspace{-4mm}
\end{figure}

OOD generalization in VQA is challenging. To generalize to out-of-distribution samples adaptively, the VQA model should own two capabilities: (1) overcoming negative language biases and (2) producing out-of-distribution answers by learning rules entailed in in-domain data. 
The prevailing OOD generalization methods~\cite{clark2019don,wu2019self,gokhale2020mutant,clark2020learning} focus on enhancing the first capability, which achieves OOD generalization by explicitly mitigating the language biases. 
While the second capability, which directly endues VQA models the potentiality to generalize to out-of-distribution (\ie, unseen or rare) samples, has not been well explored. 

Benefiting from good compositional generalization capabilities, humans can easily achieve OOD generalization~\cite{lake2017building}. 
Motivated by this, we formulate OOD generalization in VQA as a compositional generalization problem and leverage in-domain data to generate new/raw compositions of existing visual concepts (\ie, attributes and objects) to make VQA models produce out-of-distribution answers. 
As shown in Figure~\ref{fig:motivation}, given any two attribute-object pairs, such as green broccoli and brown food in Figure~\ref{fig:motivation} (a), new compositions (\eg, brown broccoli in Figure~\ref{fig:motivation} (b)) can be generated by graph generative modeling. 
The abundant recompositions probably contain unseen or rare samples in the training set, promoting the VQA models to generalize to out-of-distribution samples. 
To represent relationships and generate new compositions of existing visual concepts, Graph Generative Model~\cite{wang2018graphgan} is the natural preference, which has been proven to have the potential to improve the generalization ability of models~\cite{dai2018adversarial,zheng2020distribution}.

\begin{figure}[tbp]
\begin{center}
\includegraphics[width=1.0\linewidth]{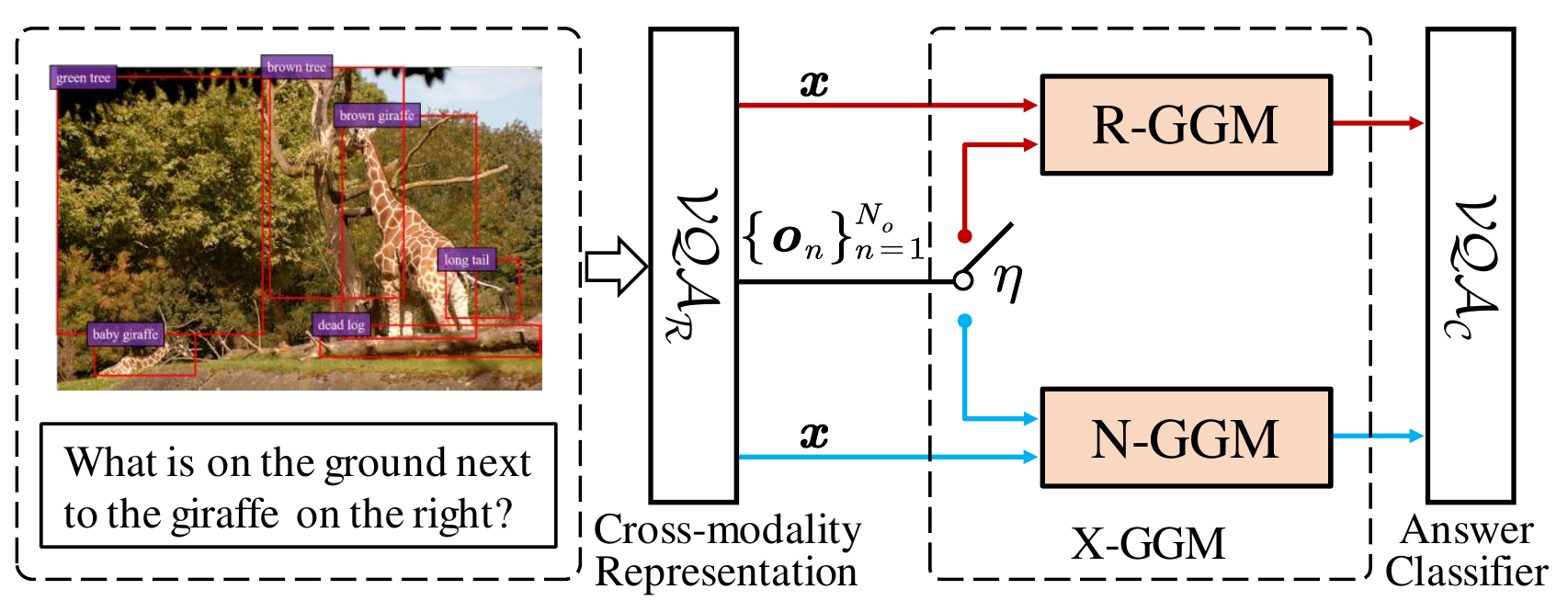}
\end{center}
\vspace{-2mm}
\caption{An overview of training a baseline VQA model using the $\ModelName$ scheme. 
$\RGenerator$ and $\NGenerator$ are the two key modules of $\ModelName$. 
$\VQAR$ and $\VQAC$ indicate the Cross-modality Representation and Answer Classifier of a baseline VQA model, respectively. 
}
\label{fig:overview}
\Description{Framework.}
\vspace{-4mm}
\end{figure}

In this paper, we propose $\ModelName$, a graph generative modeling-based training scheme, to improve the OOD generalization performance of existing VQA models while preserving their representation ability. 
As shown in Figure~\ref{fig:overview}, $\ModelName$ consists of two key modules, \ie, graph relation generative modeling ($\RGenerator$) and graph representation generative modeling ($\NGenerator$). 
$\RGenerator$ aims to update relationships between nodes to indirectly improve the OOD generalization ability. 
It first introduces adversarial perturbations (\ie, Gaussian noise) to the cross-modality representation ($\xv$) to initialize the relation matrix. 
After that, it utilizes Graph Encoder to aggregate information of different nodes and to update the relation matrix in an iterative manner. 
$\NGenerator$ targets perturbing node representations to directly enhance the OOD generalization ability. 
For $\NGenerator$, adversarial perturbations are first injected into the cross-modality representation to initialize node representations. 
Subsequently, Graph Encoder is applied to encode node representations and to update representations iteratively.

In addition, $\ModelName$ injects adversarial noises into the cross-modality representation to describe the original data distribution of relation matrices or node representations, leading to an inaccurate prior distribution in graph generative modeling. 
The inaccurate distribution affects the stability of graph adversarial learning. 
To alleviate the unstable training issue, we propose a gradient distribution consistency loss. 
The proposed loss can effectively constrain the gradient consistency between the data distribution injected into adversarial perturbations and the generated distribution (\ie, the distribution of generated relation matrices or node representations). 
Therefore, we replace the reconstruction loss in graph adversarial learning with the proposed gradient distribution consistency loss to circumvent the inaccurate prior distribution issue.

\noindent\textbf{Contribution.} Our main contributions are summarized as follows: 
($\mathbf{1}$) We propose a graph generative modeling-based training scheme, $\ModelName$, which can markedly improve the OOD generalization ability of existing VQA models while preserving their ID performance. 
($\mathbf{2}$) We propose a gradient distribution consistency loss, which can effectively alleviate the unstable gradient issue in graph adversarial learning by constraining the gradient consistency of the data distribution with adversarial perturbations and the generated distribution. 
($\mathbf{3}$) The baseline VQA model (LXMERT) trained with the proposed $\ModelName$ scheme achieves state-of-the-art OOD performance while preserving the ID performance on two standard VQA OOD benchmarks, \ie, VQA-CP v2 and GQA-OOD. 

\section{Related Work}
\label{sec:RW}

\subsection{OOD Generalization in VQA} 

Training and testing under the independent and identically distributed (i.i.d) setting have resulted in the performance of most VQA models being highly affected by superficial correlations (\ie, language biases and dataset biases)~\cite{zhang2016yin,agrawal2016analyzing,goyal2017making,agrawal2018don}. 
Recently, evaluation on the out-of-distribution (OOD) setting~\cite{guo2019quantifying,teney2020value,kervadec2020roses,gokhale2020mutant} has thus become an increasing concern for VQA. 
To improve the OOD generalization performance of VQA models, the prevailing methods target eliminating the language bias. 
Accordingly, current debiasing methods to VQA can be broadly divided into two groups, Known Bias-based~\cite{clark2019don,chen2020counterfactual,liang2020learning} and Unknown Bias-based~\cite{teney2020unshuffling,gokhale2020mutant,clark2020learning}. 

Known Bias-based methods consider the prior knowledge of language biases and design specific debiasing methods to reduce the biases explicitly existing in training set. 
Unknown Bias-based methods aim to remove the language bias or the answer distribution shift without the bias to be known in advance, which are more practical. 
In addition to considering extra human annotations~\cite{selvaraju2019taking,wu2019self,gokhale2020mutant} to balance datasets to reduce biases, most Unknown Bias-based methods leverage the adversarial training strategy~\cite{grand2019adversarial,teney2020learning,teney2020value,gokhale2020mutant,li2020closer,gan2020large}. 
Particularly, VILLA~\cite{gan2020large} introduces adversarial perturbations into embedding space and utilizes large-scale adversarial training to improve the generalization ability. 
Similarly, the proposed $\ModelName$ scheme injects adversarial perturbations into representation space and utilizes the adversarial learning strategy based on graph generative modeling to achieve OOD generalization and to mitigate language biases without prior knowledge of these biases.

\subsection{Graph Generative Models} 

Graph Generative Modeling (GGM), aiming to learn to generate new graphs with certain desirable properties by observing existing graphs, has
emerged in biomedical~\cite{de2018molgan}, social science~\cite{nauata2020house}, and graph representation learning~\cite{yu2018learning,pan2018adversarially,wang2018graphgan}. 
Recent advances in deep generative models, especially VAE~\cite{kingma2013auto} and GAN~\cite{goodfellow2014generative}, have developed towards the graph domain, inspiring many graph generative models, such as GAE~\cite{kipf2016variational}, GraphVAE~\cite{simonovsky2018graphvae}, GraphGAN~\cite{wang2018graphgan}, and Graphite~\cite{grover2019graphite}. 
These graph generative models follow the uniform framework that includes generative and discriminative models, limiting the scale of generated graphs. 
In order to handle the challenge of scalability, deep autoregressive models are utilized for graph generation, such as GraphRNN~\cite{you2018graphrnn}, GRANs~\cite{liao2019efficient}, and GraphAF~\cite{shi2020graphaf}. 
Furthermore, GRAM~\cite{kawai2019scalable} introduces graph attention mechanism to enable the graph generative model scalability. 

Recent approaches consider improving the generalization ability of graph generative models. 
You~\etal~\cite{you2020graph} propose GraphCL, a contrastive learning-based framework, to learn robust graph representations. 
Pan~\etal~\cite{pan2018adversarially} encode the topological structure and node content in a graph to obtain compact graph representations by a novel adversarial graph embedding framework. 
Dai~\etal~\cite{dai2018adversarial} propose ANE, an Adversarial Network Embedding framework, to leverage the adversarial learning principle to regularize the representation learning, contributing to the learning stable and robust graph representation. 
Zheng~\etal~\cite{zheng2020distribution} propose DB-GAN, which implicitly bridges the graph and feature spaces by prototype learning, to learn more robust graph representation. 
The proposed $\ModelName$ scheme considers generating both relationships between nodes and node representations during graph generative modeling to enhance the generalization ability of graph generative models. 


\subsection{Graph Construction in V+L Tasks}  
Graphs are non-Euclidean structured data, which can effectively represent relationships between nodes. 
Some recent works construct graphs for visual or linguistic elements in V+L tasks, such as VQA~\cite{hu2019language,li2019relation,gao2020multi}, VideoQA~\cite{huang2020location,zhu2020mucko,jiang2020reasoning}, Image Captioning~\cite{yao2018exploring,guo2019aligning,zhang2021consensus}, and Visual Grounding~\cite{jing2020visual,liu2020learning,yang2020graph}, to reveal relationships between these elements and obtain fine-grained semantic representations. 
These constructed graphs can be broadly grouped into three types: visual graphs between image objects/regions (\eg, \cite{yao2018exploring}), linguistic graphs between sentence elements/tokens (\eg, \cite{kant2020spatially}), and crossmodal graphs among visual and linguistic elements (\eg, \cite{liu2020learning}). 
In this work, we construct the visual graph for $\ModelName$.

\section{Preliminary} 
\label{sec:Preliminary} 

Taking the open-ended VQA task as a multi-class classification problem, the VQA model requires identifying the correct answer from a predefined set ($a \in \mathbb{A}$) of possible answers based on understanding the related question $q$ and image $I$. 
Commonly, a VQA model ($\VQA$) can be divided into two modules: Cross-modality Representation ($\VQAR$) and Answer Classifier ($\VQAC$),  which can be formulated by 
\begin{equation}
\begin{aligned}
\hat{a} = \VQAC (a | \xv), ~\xv = \VQAR (q, I), 
\end{aligned}
\end{equation}
where, $\VQAR$ is the crucial module of a VQA model, determining the representation ability of the VQA model. 
Therefore, the method aiming to achieve OOD generalization in VQA needs to improve the generalization performance while maintaining the representation ability of $\xv$. 
In this paper, we consider two prevailing VQA models, \ie, transformer-based model (LXMERT~\cite{tan2019lxmert}) and plain model (UpDn~\cite{anderson2018bottom}), as our baseline VQA models for evaluating the OOD performance of the proposed $\ModelName$ scheme. 

\noindentparagraph{\textbf{LXMERT.}}
LXMERT~\cite{tan2019lxmert} is a classic transformer based on cross-modality pre-training, consisting of two single-modality encoders for visual and language modalities, respectively, and one cross-modality encoder for aligning entities across the two modalities. 
It can output a visual feature sequence $\{\ov_n | \ov_n \in \mathbb{R}^d, 1 \le n \le N_o \}$, a language feature sequence, and the cross-modality feature vector $\xv \in \mathbb{R}^d$, where, $N_o$ is the number of objects in one image, and $\bm{o}_n$ is the $n$-th object feature vector. 
The visual feature sequence $\{\ov_n\}$ and the cross-modality feature vector $\xv$ will be passed into the $\ModelName$ scheme to improve the OOD generalization ability of $\xv$ by generatively modeling new relationships between objects and new node representations. 

\noindentparagraph{\textbf{UpDn.}}
Bottom-Up Top-Down model (UpDn)~\cite{anderson2018bottom} is a typical attention based VQA model. For each image $I$, it utilizes an image encoder to output the sequence of object features: $\{\ov_n | \ov_n \in \mathbb{R}^d , 1 \le n \le N_o\}$. 
For each question, it utilizes a question encoder to output the sequence of language features. 
After that, both the object and language features are fed into an attention module to obtain the joint representation $\xv \in \mathbb{R}^d$ for answer prediction. 
$\{\ov_n\}$ and $\xv$ are the inputs of the $\ModelName$ scheme.

\begin{algorithm}[!tbp]
\caption{Training a VQA Model with $\ModelName$}
\label{alg:ggm}
\begin{algorithmic}[1]
\Function {$\ModelName$}{$I, q, a, \Rmat_{GT}$}
\State $\{\ov_n, 1 \le n \le N_o \}, \xv \leftarrow \VQAR(I, q)$
\State $ cond \sim U[0, 1]$
\If {$cond \leq \eta $}  \Comment{execute $\RGenerator$}
\State $\Rmat_0 \leftarrow  \textsc{Rinit}(\xv, \sigma) $
\State $\Rmat_g, \{\bar{\vvec}_1, \dots, \bar{\vvec}_{N_k}\} \leftarrow \textsc{Rgen}(\{\ov_n\}, \Rmat_0)$ 
\State $p_{a} \leftarrow \VQAC(\xv, \bar{\vvec}_{N_k})$
\State $\mathcal{L} \leftarrow \textsc{Rlearn}(\Rmat_g, \Rmat_{GT}, p_{a})$
\Else \Comment{execute $\NGenerator$}
\State $\{\vvec_i, 1\le i\le N_o \}_0 \leftarrow  \textsc{Ninit}(\xv, \sigma) $
\State $\Vmat_g, \{\bar{\vvec}_1, \dots, \bar{\vvec}_{N_k}\} \leftarrow \textsc{Ngen}(\{\vvec_i \}_0, \Rmat_{GT})$ 
\State $p_{a} \leftarrow \VQAC(\xv, \bar{\vvec}_{N_k})$
\State $\mathcal{L} \leftarrow \textsc{Nlearn}(\Vmat_g, \{\ov_n \}, p_a)$
\EndIf
\EndFunction
\end{algorithmic}
\begin{algorithmic}[1]
\Function {$\mathrm{BMU}$}{$I, q, a$}
\State $p_{a} \leftarrow \VQA(I, q, a)$ \Comment{training baseline model}
\EndFunction
\end{algorithmic}
\end{algorithm}

\section{Graph Generative Modeling} 
\label{sec:GGM} 

To improve the OOD generalization ability of $\xv$ while preserving its representation ability, we consider the nature attribute of compositional generalization implicit in the information aggregation of graph nodes, and propose a Graph Generative Modeling-based training scheme ($\ModelName$). 
The overall process of training a baseline VQA model with the $\ModelName$ scheme is shown in \AlgRef{alg:ggm}, consisting of two parts: Graph Generative Modeling ($\ModelName$) and Baseline Model Update (BMU). 
Once $\VQAR$ yields the cross-modality representation $\xv$ and the sequence of visual feature vectors $\{\ov_n | \ov_n \in \mathbb{R}^d, 1 \le n \le N_o \}$, $\ModelName$ will update relationships between objects and representations of objects by Graph Relation Generative Modeling ($\RGenerator$) and Graph Representation Generative Modeling ($\NGenerator$) in each training update\footnote{Since the explicit relation triplets between objects are challenging to obtain, this paper operates the relation matrix and node representations instead of the adjacency matrix to achieve implicit recompositions.}. 
In the following, we first introduce the predefined object-relation graph and then detail $\RGenerator$ and $\NGenerator$. 

\subsection{Graph Construction}

For the $\ModelName$ scheme, we construct an object-relation graph using objects in one image as graph nodes. 
Specifically, the object-relation graph can be formulated as $\mathcal{G} = (\mathcal{V}, \Amat, \Rmat)$, where $\mathcal{V} = \{v_1, \dots, v_{N_o}\}$ is a set of nodes with $|\mathcal{V}|=N_o$, $\Amat$ is the adjacency matrix with $a_{i,j} \in \Amat$ marking whether there is an edge between node $v_i$ and $v_j$, and $\Rmat$ is a relation matrix describing the coherence between nodes. The constructed graph $\mathcal{G}$ is a fully-connected graph, meaning that the adjacency matrix $\Amat$ is an all-ones matrix. 
Therefore, we omit $\Amat$ in $\mathcal{G}$ for conciseness in following sections. 

To determine the ground-truth (GT) of relation matrix $\Rmat_{GT}$ utilizing in $\RGenerator$, we compute the cosine similarity between object-class embeddings $\{\cv_n | \cv_n \in \mathbb{R}^d, 1 \le n \le N_o \}$ and object-attribute embeddings $\{\av_n | \av_n \in \mathbb{R}^d, 1 \le n \le N_o \}$. These embeddings (\ie, $\cv_n$ and $\av_n$) are obtained by a pre-trained BERT model~\cite{devlin2018bert}. 
Practically, the similarity computations are random combinations of object-classes and object-attributes in one image, which will breed new combinations. 
In $\NGenerator$, the object feature $\ov_n$ yielded by $\VQAR$ serves as the GT of the $i$-th node representation (\ie, $\vvec_{GT,i} = \ov_n$). 

\subsection{Graph Relation Generative Modeling} 

Graph Relation Generative Modeling ($\RGenerator$) consists of three stages: initializing relation matrix utilizing $\xv$ and Gaussian noise (\textsc{Rinit}), Graph Relation Matrix Generation (\textsc{Rgen}), and Adversarial Learning of Relation (\textsc{Rlearn}). 

\noindentparagraph{\textbf{Initializing Relation Matrix (\textsc{Rinit}).}} 
The graph $\mathcal{G}$ is an undirected graph with $N_o$ nodes. Therefore, the relation matrix $\Rmat \in \mathbb{R}^{N_o \times N_o}$ is symmetric, meaning that only the upper triangular elements without diagonal elements (always 1) need to be initialized. 
Specifically, we first transform $\xv$ into a vector $\rv \in \mathbb{R}^{N_o(N_o-1)/2}$ by 
\begin{equation}
\begin{aligned}
\rv &= \Sigmoid \left(\Wmat_{r} \xv + \bv_{r} \right), 
\end{aligned}
\end{equation}
where, $\Wmat_r \in \mathbb{R}^{N_o(N_o-1)/2 \times d}$ is the transformation matrix, the learnable bias $\bv_{r}$ has same size as $\rv$, and $\Sigmoid$ denotes the Sigmoid activation function. 
Subsequently, we inject the adversarial perturbation (\ie, Gaussian noise) to $\rv$ and make $\hat{\rv} \sim p_{\sigma}(\hat{\rv}|\rv)$: 
\begin{equation}
\begin{aligned}
p_{\sigma}(\hat{\rv}|\rv) = \frac{1}{\sqrt{2\pi}\sigma}\exp \{-\frac{(\hat{\rv} - \rv)^2}{2\sigma^2} \}, 
\label{eq:p_sigma}
\end{aligned}
\end{equation}
where $\sigma$ is the standard deviation of Gaussian noise. 
Finally, we sequentially fill the upper triangular part ($\Rmat_0^{\bot}$) of $\Rmat_0$ with the elements of $\hat{\rv}$, and obtain the initial relation matrix by $\Rmat_0 = \Rmat_0^{\bot} + (\Rmat_0^{\bot})^{\Transpose}$. 

\begin{figure}[!t]
\centering 
\begin{center}
\includegraphics[width=\linewidth]{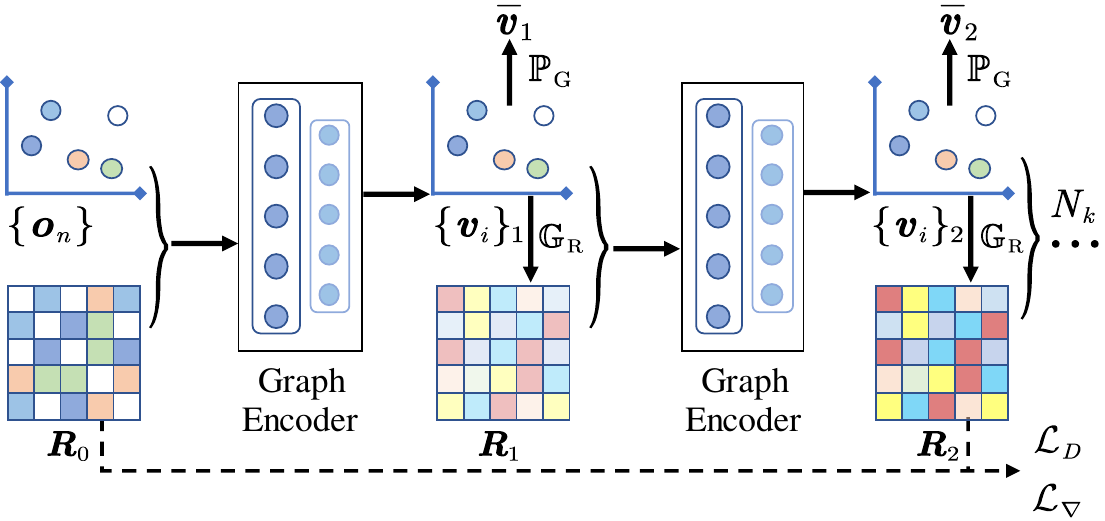}
\end{center}
\vspace{-3mm}
\caption{Graph Relation Generation. 
$\mathbb{P}_{\text{G}}$ indicates the average pooling of node representations, and $\mathbb{G}_{\text{R}}$ denotes generating relation matrix by \Eqref{eq:RG}. 
}
\Description{Architecture of \textsc{Rgen}.} 
\label{fig:Rel_GGM}
\vspace{-4mm}
\end{figure}



\noindentparagraph{\textbf{Relation Matrix Generation (\textsc{Rgen}).}}
The detailed process of \textsc{Rgen} is shown in~\Figref{fig:Rel_GGM}, which can be accomplished in $N_k$ iterations. 
Before the iteration beginning, node representations of the graph $\mathcal{G}$ are initialized by the corresponding object feature vector, \ie, $\vvec_{i}=\ov_{n}$ ($i=n$). 
In the $k$-th ($1\le k \le N_k$) iteration, 
node representations $\Vmat_{k-1} = \{\vvec_i \}_{k-1}$ 
and the relation matrix $\Rmat_{k-1}$ 
yielded by Graph Encoder in the $(k-1)$-th iteration are first fed into the $k$-th Graph Encoder and served as inputs (\ie, $\Vmat_k^{(0)}=\Vmat_{k-1} \in \mathbb{R}^{N_o \times d}$ and $\Rmat_{k}^{(0)} = \Rmat_{k-1} \in \mathbb{R}^{N_o \times N_o}$). 
Subsequently, the node representations are encoded by Graph Encoder, which is essentially a GCN~\cite{kipf2016semi} with $N_l$-layers. 
Specifically, the node representations in the $l$-layer of Graph Encoder can be updated with 
\begin{equation}
\Vmat^{(l+1)}_k = \Sigmoid((\Rmat_{k}^{(0)}\Vmat^{(l)}_k)\Wmat^{(l)}_k + \bv^{(l)}_k), 
\end{equation}
where, $\Wmat^l_k \in \mathbb{R}^{N_o \times d \times d}$ is spanned by $N_o$ matrices of feature transformation, and $\bv_k^l \in \mathbb{R}^{N_o \times d}$ is the learnable bias for the transformation. 
After that, we assemble the input and the output of each layer in Graph Encoder to obtain the final output $\Vmat_{k}$ of the $k$-th iteration: 
\begin{equation}
\Vmat_{k} = \sum_{l=0}^{N_l} \ReLU(\Vmat_k^{(l)}\Wmat_a + \bv_a), 
\label{eq:Encoder_sum}
\end{equation} 
where, $\ReLU$ denotes the ReLU activation function, $\Wmat_a \in \mathbb{R}^{N_o \times d \times d}$ and $\bv_a \in \mathbb{R}^{N_o \times d}$ are the weight and bias for node feature transformation, respectively. 
Finally, the relation matrix is generated using the final node representations $\Vmat_{k}$ by 
\begin{equation}
\Rmat_k = \Sigmoid(\Vmat_k\Vmat_k^{\Transpose}).
\label{eq:RG}
\end{equation}
We utilize the output of the last iteration as the final generated relation matrix, that is, $\Rmat_g = \Rmat_{N_k}$. 
In addition, we average node representations $\{\vvec_i, 1\le i \le N_o\}_{k}$ to derive the graph representation $\bar{\vvec}_{k}$ in the $k$-th iteration. 

\begin{figure}[!t]
\centering 
\begin{center}
\includegraphics[width=\linewidth]{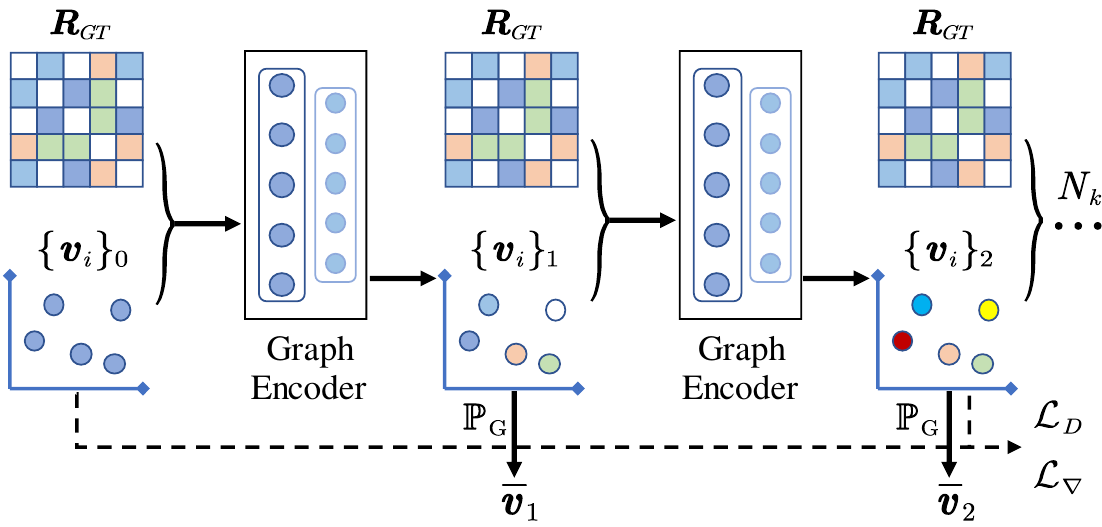}
\end{center}
\vspace{-3mm}
\caption{Graph Representation Generation. 
$\mathbb{P}_{\text{G}}$ is an average operation of node representations. 
}
\Description{Architecture of \textsc{Ngen}.} 
\label{fig:Rep_GGM}
\vspace{-4mm}
\end{figure}

\noindentparagraph{\textbf{Adversarial Learning of $\RGenerator$ (\textsc{Rlearn}).}} 
When initializing the relation matrix in the stage of \textsc{Rinit}, we find that the process of $\RGenerator$ introducing adversarial perturbations is similar to that of Denoising AutoEncoder~\cite{vincent2008extracting}. 
From the works~\cite{li2019learning,bigdeli2020learning}, we know that the optimal Denoising AutoEncoder for additive white Gaussian noise (AWGN) can be explicitly calculated, which is related to the gradient of the data distribution after adding noises.
Practically, the logarithmic gradient of the data distribution with adversarial perturbations can be deduced from~\Eqref{eq:p_sigma}: 
\begin{equation}
\nabla_{\hat{\rv}}\log p_{\sigma}(\hat{\rv}|\rv) = -\frac{(\hat{\rv} - \rv)^2}{\sigma^2}. 
\end{equation} 
After obtaining $\nabla_{\hat{\rv}}\log p_{\sigma}(\hat{\rv}|\rv)$, 
we can compute the gradient of log-likelihood estimation, \ie, $\nabla_{\rv}\log p_g(\rv)$, of elements in the generated relation matrix $\Rmat_g$
to constrain the gradient distribution consistency between the generated relation matrix $\Rmat_g$ and the pre-derived similarity matrix $\Rmat_{GT}$. 
Therefore, the gradient distribution consistency loss can be defined as 
\begin{equation} 
\mathcal{L}_{\nabla} \triangleq  \mathbb{E}\left[\left\| \nabla_{\rv}\log p_g(\rv) - \nabla_{\hat{\rv}}\log p_{\sigma}(\hat{\rv}|\rv) \right\|_2^2\right]. 
\label{eq:L_G}
\end{equation} 

In addition to constraining the gradient distribution consistency, 
we further utilize KL-Divergence to measure the distance between the generated distribution $p_g(\rv)$ and the raw data distribution $p_{data}(\rv)$, where $p_g$ and $p_{data}$ denote probability distributions of elements in matrix $\Rmat_g$ and $\Rmat_{GT}$, respectively. 
The distance measure loss of training our $\RGenerator$ can be formulated by 
\begin{equation}
\mathcal{L}_D = \mathrm{KL}(p_g \| p_{data}) + \mathrm{KL}(p_{data} \| p_g), 
\label{eq:L_D}
\end{equation}
where, $\mathrm{KL}(p\|q)=\mathbb{E}_{\rv \sim p}\left[\log p(\rv) - \log q(\rv)\right]$.

Besides the adversarial losses (\ie, $\mathcal{L}_{\nabla}$ and $\mathcal{L}_{D}$), we compute the BCE loss ($\mathcal{L}_{\textsc{bec}}$) of the VQA task by adding $\xv$ (the original cross-modality representation) and $\bar{\vvec}_{N_k}$ (the graph representation yielded by the last iteration in \textsc{Rgen}) and passing the value into $\VQAC$ to derive the likelihood estimation of answer distribution $p_{a}$. 

In summary, the total loss of training $\RGenerator$ can be formulated by 
\begin{equation}
\mathcal{L} = \alpha \mathcal{L}_{\nabla} + \beta \mathcal{L}_D + \mathcal{L}_{\textsc{bce}}, 
\label{eq:L_total}
\end{equation} 
where, $\alpha$ and $\beta$ are weighted coefficients. 

\begin{table*}[!tbp]
\begin{center}
\caption{Statistics of VQA-CP v2 and GQA-OOD.}
\vspace{-3mm}
\setlength{\tabcolsep}{2.0mm}{
\centering 
\renewcommand\arraystretch{1.1}
\begin{tabular}{l|ccccc|ccccc|c}
\hline
\multirow{2}{*}{Dataset}  
&\multicolumn{5}{c|}{Image} 
&\multicolumn{5}{c|}{Question} 
&\multirow{2}{*}{Year} 
\\
\cline{2-11}
&Train &Val &OOD Test &ID Test &Total
&Train &Val &OOD Test &ID Test &Total
& 
\\ 
\hline
\hline
VQA-CP v2~\cite{agrawal2018don}  
& \eqmakebox[log][r]{119,107} & \eqmakebox[log][r]{31,976} 
& \eqmakebox[log][r]{89,437} & \eqmakebox[log][r]{32,266} 
& \eqmakebox[log][r]{272,786} 
& \eqmakebox[log][r]{388,735} & \eqmakebox[log][r]{38,802}
& \eqmakebox[log][r]{179,928} & \eqmakebox[log][r]{40,000} 
& \eqmakebox[log][r]{647,465} 
&2018 
\\
GQA-OOD~\cite{kervadec2020roses}  
& \eqmakebox[log][r]{72,140} & \eqmakebox[log][r]{9,406} 
& \eqmakebox[log][r]{330} & \eqmakebox[log][r]{365} 
& \eqmakebox[log][r]{82,241} 
& \eqmakebox[log][r]{943,000} & \eqmakebox[log][r]{51,045} 
& \eqmakebox[log][r]{1,063} & \eqmakebox[log][r]{1,733} 
& \eqmakebox[log][r]{996,841} 
&2020 
\\ 
\hline
\end{tabular}
}
\label{tab:dataset}
\end{center}
\vspace{-1mm}
\end{table*}

\subsection{Graph Representation Generative Modeling}

Graph Representation Generative Modeling ($\NGenerator$) is similar to $\RGenerator$, which can also be implemented in three stages: 
initializing node representations with $\xv$ and Gaussian noise (\textsc{Ninit}), Graph Node Representation Generation (\textsc{Ngen}), and Adversarial Learning of Node Representation (\textsc{Nlearn}).

\noindentparagraph{\textbf{Initializing Node Representations (\textsc{Ninit}).}} 
To obtain the $N_o$ initial node representations of the graph $\mathcal{G}$, we first span $\xv$ to $[\xv, \dots, \xv] \in \mathbb{R}^{N_o\times d}$, and utilize a fully-connected layer ($\mathcal{M}_{\text{FC}}$) to increase the variability of node representations: 
\begin{equation}
\begin{aligned}
\Vmat_{\xv} &= \mathcal{M}_{\text{FC}} ([\xv, \dots, \xv]), \Vmat_{\xv} \in \mathbb{R}^{N_o\times d},  
\end{aligned}
\end{equation}
where, $\Vmat_{\xv}=\{\vvec_{i,\xv} | \vvec_{i,\xv} \in \mathbb{R}^{d} \}, 1\le i\le N_o$. 
After that, we inject adversarial perturbations to $\Vmat_{\xv}$, and make the sampled feature vector $\hat{\vvec}_{i, \xv}$ of the $i$-th element in $\Vmat_{\xv}$ satisfy the noise prior distribution $p_{\sigma}(\hat{\vvec}_{i, \xv} | \vvec_{i, \xv})$: 
\begin{equation}
\begin{aligned}
p_{\sigma}(\hat{\vvec}_{i, \xv} | \vvec_{i, \xv}) = \frac{1}{\sqrt{2\pi}\sigma}\exp \{-\frac{(\hat{\vvec}_{i, \xv} - \vvec_{i, \xv})^2}{2\sigma^2} \}. 
\end{aligned}
\end{equation}
Finally, the initial node representations of the graph $\mathcal{G}$ can be obtained by $\Vmat_0 = \hat{\Vmat}_{\xv} = [\hat{\vvec}_{0,\xv}, \dots, \hat{\vvec}_{N_o,\xv}]$. 

\noindentparagraph{\textbf{Node Representation Generation (\textsc{Ngen}).}} 
The details of \textsc{Ngen} are shown in Figure~\ref{fig:Rep_GGM}. 
In the $k$-th iteration, the node representations $\Vmat_{k-1}= \{\vvec_i, 1 \le i \le N_o \}_{k-1}$ generated by Graph Encoder in the $(k-1)$-th iteration and the predefined GT relation matrix $\Rmat_{GT}$ are first passed into the $k$-th Graph Encoder to update the node representations of each layer. 
After that, the final output of the $k$-th Graph Encoder $\Vmat_k$ can be obtained by~\Eqref{eq:Encoder_sum}. 
The architecture of Graph Encoder in $\NGenerator$ is the same as that in $\RGenerator$. 
We utilize the output of the last iteration as the generated node representations, \ie, $\Vmat_g = \Vmat_{N_k}$.
Besides, in each iteration, we average $N_o$ node representations and yield the graph representation $\bar{\vvec}_k$.

\noindentparagraph{\textbf{Adversarial Learning of $\NGenerator$ (\textsc{Nlearn}).}} 
The total loss $\mathcal{L}$ of training $\NGenerator$ is also a weighted sum of $\mathcal{L}_{\nabla}$, $\mathcal{L}_D$, and $\mathcal{L}_{\textsc{bce}}$. 
Specifically, the gradient distribution consistency loss $\mathcal{L}_{\nabla}$ is defined as the distance between the distribution gradient of the initial node representations after adding Gaussian noise ($\hat{\Vmat}_{\xv}$) and the log-likelihood estimation of generated node representations ($\Vmat_g$). 
$\mathcal{L}_D$ is the KL-Divergence loss between the distribution of generated node representations ($\Vmat_g$) and the distribution of the raw node representations ($\{\ov_n\}$) yielded by $\VQAR$. 
The VQA classification loss $\mathcal{L}_{\textsc{bce}}$ is the same as the corresponding term in~\Eqref{eq:L_total}.

\section{Experiments}
\label{sec:Experiments}

In this section, we evaluate the effectiveness of $\ModelName$ key components and compare the proposed $\ModelName$ scheme with state-of-the-art approaches on two standard VQA OOD benchmarks, \ie, VQA-CP v2~\cite{agrawal2018don} and GQA-OOD~\cite{kervadec2020roses}. 
Table~\ref{tab:dataset} reports the statistics of experimental datasets. 

\subsection{Experimental Settings}

\begin{table}[!tbp]
\begin{center}
\caption{Ablation studies on $\RGenerator$ and $\NGenerator$ with different combinations of $\mathcal{L}_D$ and $\mathcal{L}_{\nabla}$ (More results are shown in Table~\ref{tab:abl_comps_supp} of the Appendix). 
\textcolor{blue}{$\uparrow$} (\textcolor{red}{$\downarrow$}) indicates the higher (lower) the better. LX. means the Baseline, \ie, LXMERT. 
}
\vspace{-2mm}
\label{tab:abl_comps}
\centering
\setlength{\tabcolsep}{0.95mm}{
\renewcommand\arraystretch{1.1}
\begin{tabular}{c|cc|cc|ccc|ccc}
\hline
\multirow{2}{*}{S/N}
&\multicolumn{2}{c|}{$\RGenerator$}
&\multicolumn{2}{c|}{$\NGenerator$} 
&\multicolumn{3}{c|}{VQA-CP v2}
&\multicolumn{3}{c}{GQA-OOD} 
\\
\cline{2-11}
&\small $\mathcal{L}_{\nabla}$ 
&\small $\mathcal{L}_D$ 
&\small $\mathcal{L}_{\nabla}$
&\small $\mathcal{L}_D$ 
&\small OOD\textcolor{blue}{$\uparrow$}
&\small ID\textcolor{blue}{$\uparrow$}
&\small Gap\textcolor{red}{$\downarrow$} 
&\small Tail\textcolor{blue}{$\uparrow$} 
&\small Head\textcolor{blue}{$\uparrow$} 
&$\Delta$\textcolor{red}{$\downarrow$}
\\
\hline
\hline
\rowcolor{gray!15}\small LX. 
&\multicolumn{4}{c|}{---} 
&59.98 &65.21 &5.23 
&49.80 &57.70 &15.90 \\ 
\hline 
\#1 &\checkmark &\checkmark & & 
&63.84 &66.41 &2.57 
&51.39 &56.93 &10.77 
\\
\#2 & & &\checkmark &\checkmark 
&62.90 &65.57 &2.67
&51.27 &57.30 &11.76
\\
\hline 
\#3 &\checkmark & &\checkmark & 
&63.75 &66.03 &2.28 
&51.65 &57.13 &10.61 
\\ 
\#4 & &\checkmark & &\checkmark 
&62.73 &65.99 &3.26 
&51.27 &57.19 &11.54 
\\
\hline 
\#5 &\checkmark &\checkmark &\checkmark &\checkmark 
&64.33 &66.92 &2.59 
&52.47 &57.51 &9.59 
\\
\hline
\end{tabular}
} 
\end{center}
\end{table}

\subsubsection{Evaluation Benchmarks} 

\noindentparagraph{\textbf{VQA-CP v2.}}
It is the reorganization of the training and validation sets of VQA v2.0~\cite{shih2016look}. 
To evaluate the OOD generalization performance of existing VQA models, Agrawal~\etal~\cite{agrawal2018don} construct VQA-CP v2 by ensuring that the answer distribution of per question type in the test set and the training set is different. 
The original VQA-CP v2 only consists of the test set (about 98K images, 220K questions, and 2.2M answers) and the training set (about 121K images, 438K questions, and 4.4M answers). That is, the benchmark does not have an official validation set. 

\noindentparagraph{\textbf{GQA-OOD.}} 
It is a fine-grained restructuring of GQA~\cite{hudson2019gqa}, which obtains the validation and test sets by introducing answer distribution shifts into the validation and test sets of GQA. 
However, GQA-OOD shares the same training set with GQA. 
The GQA-OOD benchmark provides four metrics. 1) \textbf{Tail}: accuracy on OOD samples (\ie, samples of the tail of the answer class distribution), 2) \textbf{Head}: accuracy on in-domain (ID) samples (\ie, samples from the head distribution, 3) \textbf{All}: overall accuracy on all test samples, and 4) $\Delta (\%)\textup{ = (head - tail) / tail}$: a new evaluation protocol proposed by Kervadec~\etal~\cite{kervadec2020roses} to illustrate how much is the error prediction imbalanced between frequent and rare answers. 

\subsubsection{Experimental Details}

\noindentparagraph{\textbf{Data Preprocessing.}} 
As Teney~\etal~\cite{teney2020value} discussed, there are several pitfalls in evaluating VQA models under the OOD setting on VQA-CP v2, especially using the OOD test set for model selection and evaluating the ID performance of a VQA model after retraining the model on VQA v2.0~\cite{shih2016look}. 
Therefore, following recent works~\cite{teney2019actively,teney2020unshuffling,teney2020learning,clark2020learning}, we re-split VQA-CP v2. 
Specifically, we reserve 40k random samples from the train set to evaluate the in-domain (ID) performance and 40k random instances from the test set for model selection. 
Besides, the bottom-up attention Faster R-CNN~\cite{anderson2018bottom} pre-trained on Visual Genome~\cite{krishna2017visual} is utilized to obtain object class and attribute in the image for the two datasets. 

\noindentparagraph{\textbf{Implementation Details.}}
For $\ModelName$, we set the number $N_o$ of nodes in $\mathcal{G}$ to 36 and the standard feature dimensionality $d$ to 768. 
The standard deviation $\sigma$ in \Eqref{eq:p_sigma} is 1.0. 
The threshold $\eta$ in Algorithm~\ref{alg:ggm} are 0.8 and 0.5 for VQA-CP v2 and GQA-OOD, respectively. 
The total step $N_k$ of $\ModelName$ is 2, and the layer $N_l$ of the encoder in each step is 2. 
The weight coefficients $\alpha$ and $\beta$ in \Eqref{eq:L_total} are 6 and 72 for $\RGenerator$, and 6.6 and 0.17 for $\NGenerator$. 
All experiments are implemented on one NVIDIA GTX2080 12GB GPU for 5 epochs with a batch size of 128 and a learning rate of 1e-6. 

\subsection{Ablation Studies} 

We conduct ablation studies on key components of $\ModelName$ using LXMERT as the baseline VQA model\footnote{Pre-trained transformers usually have smaller ID/OOD generalization gaps than previous plain models~\cite{hendrycks2020pretrained}. Considering that improving the OOD performance of transformer-based methods is more difficult than plain methods, we utilize a transformer-based model as the baseline to conduct our ablation studies.}. 
Considering the discovery in work~\cite{teney2020value}, \ie, the questions with yes/no/number answers on VQA-CP v2 are easier to game, for a fairer comparison, we train all models on VQA-CP v2 ``All''. But, we select the best model and focus the model analysis on VQA-CP v2 ``other''.

\subsubsection{Effectiveness of Key Components}
\noindentparagraph{\textbf{$\RGenerator$ / $\NGenerator$ / $\ModelName$} \textit{vs.} \textbf{Baseline.}} 
To improve the OOD generalization ability of the cross-modality representation $\xv$ yielded by $\VQAR$, besides training the baseline VQA model using the full $\ModelName$ scheme, $\RGenerator$ and $\NGenerator$ can also work independently. 
Therefore, we consider comparisons between the ablated components (\#1, \#2, \#5) and the Baseline, \ie, LXMERT (LX.) in Table~\ref{tab:abl_comps} to evaluate the respective effects of $\RGenerator$, $\NGenerator$, and $\ModelName$. 
Specifically, $\blacktriangleright$ LX. \vs \#1 (only using $\RGenerator$ in the training process). 
$\blacktriangleright$ LX. \vs \#2 (only using $\NGenerator$ in the training process). 
$\blacktriangleright$ LX. \vs \#5 (using the full $\ModelName$ scheme that executes $\RGenerator$ and $\NGenerator$ at a ratio of $\eta$ in the training process). 
Table~\ref{tab:abl_comps} shows the results on VQA-CP v2 ``Other'' and GQA-OOD. 
LX. \vs \#5 demonstrates the effectiveness of $\ModelName$. 
LX. \vs \#1 / \#2 illustrates that the single $\RGenerator$ / $\NGenerator$ can also effectively improve the OOD generalization performance of the baseline VQA model. 

\noindentparagraph{\textbf{$\RGenerator$} / \textbf{$\NGenerator$} \textit{vs.} \textbf{$\ModelName$.}} 
In each training iteration, $\ModelName$ randomly executes $\RGenerator$ or $\NGenerator$ to improve the OOD generalization ability of the baseline VQA model. 
To explore the importance of $\RGenerator$ and $\NGenerator$ to the full training scheme ($\ModelName$), as shown in Table~\ref{tab:abl_comps} (\#1, \#2, \#5), we conduct the following comparisons: 
$\blacktriangleright$ \#1 \vs \#5, 
and $\blacktriangleright$ \#2 \vs \#5. 
The results in Table~\ref{tab:abl_comps} indicate that $\RGenerator$ is more effective than $\NGenerator$ for the $\ModelName$ scheme in boosting the OOD performance of the baseline VQA model (LXMERT) on both VQA-CP v2 ``Other'' and GQA-OOD. 
This is mainly because the gradient instability caused by the noise introduction in $\NGenerator$ is larger than that in $\RGenerator$. 
Specifically, the noise vector dimensions introduced into $\RGenerator$ and $\NGenerator$ are $N_o (N_o - 1)/2 = 630$ and $N_o \times d = 27,648$, respectively.

\noindentparagraph{\textbf{Gradient Distribution Consistency Loss.}} 
To evaluate the effectiveness of the gradient distribution consistency loss $\mathcal{L}_{\nabla}$ in alleviating the unstable gradient issue, we compare the respective effects of $\mathcal{L}_{\nabla}$ and $\mathcal{L}_D$. 
The KL-Divergence loss $\mathcal{L}_D$ is also an adversarial loss. 
The experimental settings are shown in Table~\ref{tab:abl_comps} (\#3, \#4, \#5). 
More specifically, 
$\vartriangleright$ \#$\mathsf{3}$: only using $\mathcal{L}_D$ as the adversarial loss to train the baseline VQA model (LXMERT) with the $\ModelName$ scheme, 
$\vartriangleright$ \#$\mathsf{4}$: only using $\mathcal{L}_{\nabla}$ as the adversarial loss to train the baseline VQA model with the $\ModelName$ scheme, 
and $\vartriangleright$ \#$\mathsf{5}$: using a weighted sum of $\mathcal{L}_{\nabla}$ and $\mathcal{L}_D$ as the adversarial loss to train the baseline VQA model with the $\ModelName$ scheme. 
The comparisons between \#3, \#4, and \#5 on VQA-CP v2 ``Other'' and GQA-OOD in Table~\ref{tab:abl_comps} illustrate that both $\mathcal{L}_D$ and $\mathcal{L}_{\nabla}$ are effective for training the baseline VQA model with the $\ModelName$ scheme. 
However, $\mathcal{L}_{\nabla}$ performs better than $\mathcal{L}_D$, demonstrating the effectiveness of $\mathcal{L}_{\nabla}$ in alleviating unstable gradient updates in graph adversarial learning. 

\begin{table}[!t]
\begin{center}
\caption{Comparisons between different variants of Graph Encoder (GEnc.) with $N_l$ layers. 
}
\vspace{-2mm}
\label{tab:abl_GEnc}
\setlength{\tabcolsep}{1.9mm}{
\centering 
\begin{tabular}{l|c|ccc|ccc}
\hline
\multirow{2}{*}{GEnc.}
&\multirow{2}{*}{$N_l$}
&\multicolumn{3}{c|}{VQA-CP v2} 
&\multicolumn{3}{c}{GQA-OOD} \\
\cline{3-8}
& 
&\small OOD\textcolor{blue}{$\uparrow$}
&\small ID\textcolor{blue}{$\uparrow$}
&\small Gap\textcolor{red}{$\downarrow$} 
&\small Tail\textcolor{blue}{$\uparrow$} 
&\small Head\textcolor{blue}{$\uparrow$} 
&$\Delta$\textcolor{red}{$\downarrow$} \\
\hline
\hline
\rowcolor{gray!15}LX. &--- 
&59.98 &65.21 &5.23 
&49.80 &57.70 &15.90 \\
\hline
\multirow{2}{*}{GCN} &1 
&64.10 &66.70 &2.60
&51.65 &57.24 &10.83 \\
~&2 
&\textbf{64.33} &\textbf{66.92} &2.59 
&\textbf{52.47} &{57.51} &\textbf{9.59} \\
\hline
\multirow{2}{*}{GIN} &1 
&63.62 &66.24 &2.62 
&50.99 &57.08 &11.93 \\
~&2 
&63.83 &66.39 &2.56 
&51.27 &57.30 &11.76 \\
\hline
\multirow{2}{*}{GAT} &1 
&63.94 &66.40 &\textbf{2.46} 
&51.65 &\textbf{57.71} &11.73 
\\
~&2 
&63.94 &66.45 &2.51 
&50.99 &56.90 &11.59
\\
\hline
\end{tabular}
} 
\end{center}
\end{table}

\begin{table*}[!t]
\begin{center}
\caption{
Comparisons with UB-based debiasing methods on VQA-CP v2 (full comparisons with state-of-the-arts are reported in Table~\ref{tab:vqacpv2_SoTA_full} of the Appendix). 
$^{\dagger}$: methods using additional manually annotated data, $^{*}$: results from our reimplementation using the official released code, ---: not reported.
}
\vspace{-2mm}
\label{tab:vqacpv2}
\centering 
\setlength{\tabcolsep}{1.8mm}{
\begin{tabular}{|l|cccc|cccc|cccc|}
\hline 
\multirow{2}{*}{Method} 
&\multicolumn{4}{c|}{VQA-CP v2 (OOD Test) \textcolor{blue}{$\uparrow$}} 
&\multicolumn{4}{c|}{VQA-CP v2 (ID Test) \textcolor{blue}{$\uparrow$}} 
&\multicolumn{4}{c|}{Gap \textcolor{red}{$\downarrow$}} \\ 
\cline{2-13}
&All &Yes/No &Num. &Other 
&All &Yes/No &Num. &Other 
&All &Yes/No &Num. &Other \\ 
\hline 
\hline
Unshuffling~\cite{teney2020unshuffling} 
&42.39 &47.72 &14.43 &\cellcolor{cyan!10}{47.24} 
&--- &--- &--- &\cellcolor{cyan!10}{---} 
&--- &--- &--- &\cellcolor{cyan!10}{---} \\
Unshuffling+CF~\cite{teney2020learning} 
&40.60 &61.30 &15.60 &\cellcolor{cyan!10}{46.00} 
&63.30 &79.40 &45.50 &\cellcolor{cyan!10}{53.70} 
&22.70 &18.10 &29.90 &\cellcolor{cyan!10}{7.70} \\
Unshuffling+CF+GS~\cite{teney2020learning}
&46.80 &64.50 &15.30 &\cellcolor{cyan!10}{45.90}
&62.40 &77.80 &43.80 &\cellcolor{cyan!10}{53.60}
&15.60 &13.30 &28.50 &\cellcolor{cyan!10}{7.70} \\ 

\hline
UpDn~\cite{anderson2018bottom} 
&38.82  &42.98  &12.18  &\cellcolor{cyan!10}{43.95} 
&64.73  &79.45  &49.59  &\cellcolor{cyan!10}{55.66}  
&25.91  &36.47 &37.32 &\cellcolor{cyan!10}{11.71} \\
\quad + Top answer masked~\cite{teney2020value} 
&40.61  &82.44  &27.63  &\cellcolor{cyan!10}{22.26} 
&30.90  &44.12  &5.00  &\cellcolor{cyan!10}{20.85} 
&9.71 &38.32 &22.63 &\cellcolor{cyan!10}{{\textbf{1.41}}} \\ 
\quad + AdvReg~\cite{grand2019adversarial} 
&36.33  &59.33  &14.01  &\cellcolor{cyan!10}{30.41} 
&50.63  &67.39  &38.81  &\cellcolor{cyan!10}{38.37} 
&14.30 &8.06 &24.80 &\cellcolor{cyan!10}{7.96} \\ 
\quad + GRL~\cite{grand2019adversarial}
&42.33 &59.74 &14.78 &\cellcolor{cyan!10}{40.76} 
&56.90 &69.23 &42.50 &\cellcolor{cyan!10}{49.36} 
&14.57 &9.49 &27.72 &\cellcolor{cyan!10}{8.60} \\
\quad + RandImg ($\lambda$=12)~\cite{teney2020value} 
&55.37  &83.89  &41.60  &\cellcolor{cyan!10}{44.20}  
&54.24  &64.22  &34.40  &\cellcolor{cyan!10}{50.46}
&\text{1.13} &19.67 &7.20 &\cellcolor{cyan!10}{6.26} \\
\quad + RandImg ($\lambda$=5)~\cite{teney2020value}
&51.15  &75.06  &24.30  &\cellcolor{cyan!10}{45.99}  
&59.28  &70.66  &43.06  &\cellcolor{cyan!10}{53.40} 
&{8.13} &4.40 &18.76 &\cellcolor{cyan!10}{7.41} \\ 
\quad + MUTANT$^{\dagger}$~\cite{gokhale2020mutant}
&61.72 &88.90 &49.68 &\cellcolor{cyan!10}{50.78} 
&--- &--- &--- &\cellcolor{cyan!10}{---} 
&--- &--- &--- &\cellcolor{cyan!10}{---} \\
\quad + $\ModelName$ 
&45.71  &43.48  &27.65  &\cellcolor{cyan!10}{{52.34}}   
&67.16  &83.74  &48.26  &\cellcolor{cyan!10}{{56.91}} 
&21.45  &40.26  &20.61  &\cellcolor{cyan!10}{\text{4.57}} \\ 
\hline

\hline
LXMERT$^{*}$~\cite{tan2019lxmert} 
&63.90 &80.45 &46.58 &\cellcolor{cyan!10}{{59.98}} 
&75.57 &91.39 &59.83 &\cellcolor{cyan!10}{65.21}
&11.67 &10.94 &13.25 &\cellcolor{cyan!10}{5.23} \\

\quad + MUTANT$^{\dagger}$~\cite{gokhale2020mutant}
&{69.52} &93.15 &67.17 &\cellcolor{cyan!10}{57.78} 
&--- &--- &--- &\cellcolor{cyan!10}{---} 
&--- &--- &--- &\cellcolor{cyan!10}{---} \\

\quad + VILLA$^*$~\cite{gan2020large} 
&48.06 &42.66 &18.97 &\cellcolor{cyan!10}{58.89} 
&{75.85} &90.40 &60.02 &\cellcolor{cyan!10}{{66.66}} 
&27.79 &47.74 &41.05 &\cellcolor{cyan!10}{7.77} \\
\quad + $\ModelName$ 
&66.95 &78.91 &53.63 &\cellcolor{cyan!10}{{\textbf{64.33}}} 
&75.88 &90.18 &60.06 &\cellcolor{cyan!10}{{\textbf{66.92}}} 
&8.93 &11.27 &6.43 &\cellcolor{cyan!10}{{{2.59}}} \\
\hline 
\end{tabular}
}
\end{center}
\end{table*}

\subsubsection{Variants of Graph Encoder} 

In the proposed $\ModelName$ scheme, Graph Encoder is a general block utilized to aggregate node information and encode relationships between nodes. 
That is, architecture differences between different Graph Encoders can not markedly affect the performance of~$\ModelName$ to OOD generalization. 
To demonstrate that the $\ModelName$ scheme is robust for different variants of Graph Encoder, we consider three prevailing Graph Neural Networks (\ie, GCN~\cite{kipf2016semi}, GIN~\cite{xu2018powerful}, and GAT~\cite{velivckovic2017graph}) and set the number of their layers $N_l$ to 1 and 2.  
From the results on  VQA-CP v2 ``Other'' and GQA-OOD shown in Table~\ref{tab:abl_GEnc}, we can observe that there is only a slight difference between the OOD performance of different Graph Encoder variants. 
However, all of these variants have achieved dramatic improvements in OOD performance compared with the baseline VQA model (LXMERT). 
The results demonstrate that $\ModelName$ is agnostic to Graph Encoder and is robust for different Graph Encoders. 

\subsubsection{Hyperparameter} 
\begin{figure}[!tbp]
\begin{center}
\includegraphics[width=1.0\linewidth]{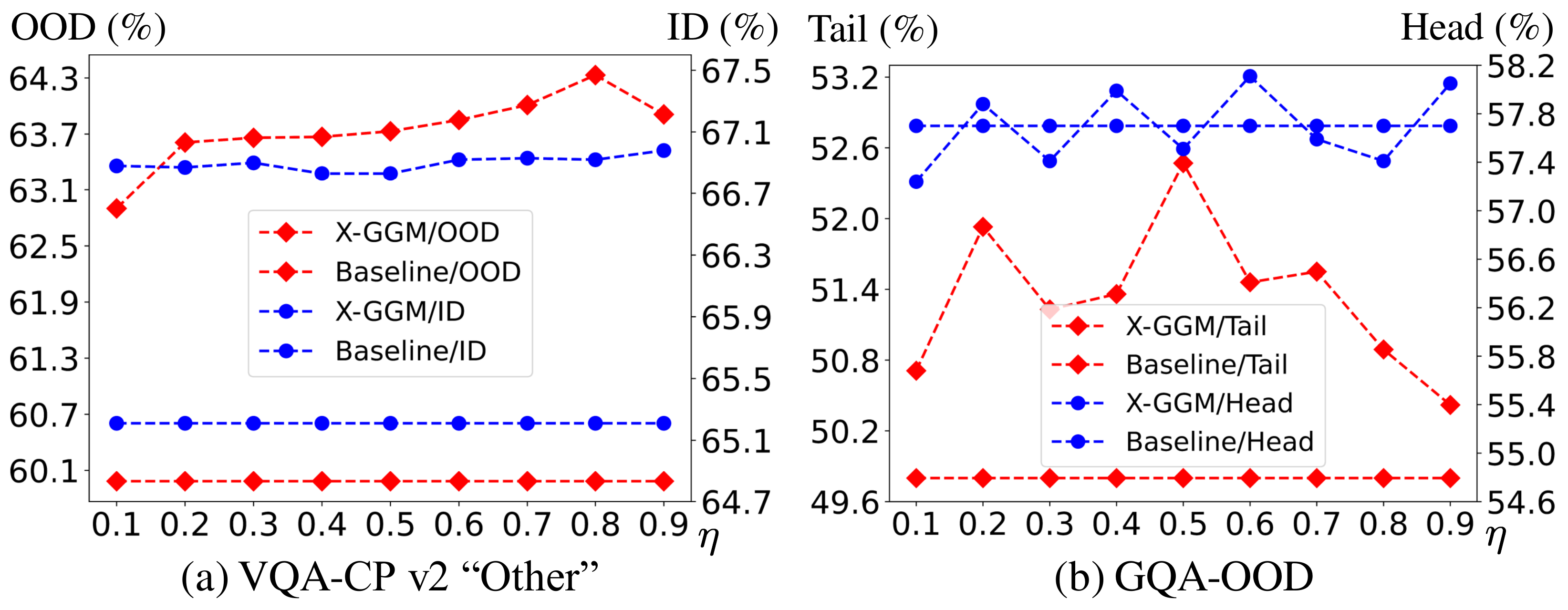}
\end{center}
\vspace{-3mm}
\caption{Ablation studies on the hyperparameter $\eta$. 
(a) OOD and ID accuracies of our method ($\ModelName$) and the Baseline (LXMERT) on VQA-CP v2 ``Other''. 
(b) OOD (Tail) and ID (Head) accuracies on GQA-OOD. 
}
\label{fig:abl_params}
\Description{Hyparameter.}
\vspace{-2mm}
\end{figure}

\noindentparagraph{\textbf{Ratio of $\RGenerator$ and $\NGenerator$ $\eta$.}} The threshold $\eta$ in \AlgRef{alg:ggm} determines the ratio of $\RGenerator$ and $\NGenerator$ executed in each training iteration. 
To analyze the influence of $\eta$ for $\ModelName$ improving the OOD performance, we conduct some ablated experiments on different values of $\eta$, \ie, $\eta \in \{0.1, 0.2, \dots, 0.9\}$. 
Results on VQA-CP v2 ``Other'' and GQA-OOD are shown in Figure~\ref{fig:abl_params} (a) and (b), respectively. 
We observe that the baseline VQA model (LXMERT) trained with the proposed $\ModelName$ scheme can achieve the best OOD performance on VQA-CP v2 ``Other'' and GQA-OOD when $\eta$ = 0.8 and $\eta$ = 0.5, respectively. 
The average gains of the OOD performance on VQA-CP v2 ``Other'' and GQA-OOD are $+3.76_{\pm 0.36}$ and $+1.54_{\pm 0.59}$, respectively. 
Comparing the OOD performance gains (\wrt the baseline VQA model) with the standard deviations of the OOD performance, we can consider the performance fluctuation is slight, demonstrating the robustness of the $\ModelName$ scheme to the hyperparameter $\eta$.

\subsection{Comparisons with State-of-the-Arts} 

To evaluate the OOD and ID performance of baseline VQA models trained with the proposed $\ModelName$ scheme, we compare them with state-of-the-art methods on VQA-CP v2 and GQA-OOD.

\subsubsection{Performance on VQA-CP v2} 

\noindentparagraph{\textbf{Setting.}} 
$\ModelName$ is a training scheme for improving the OOD generalization ability of baseline VQA models while preserving the ID performance, which alleviates the language bias in VQA to some extent without the bias being known in advance. 
Therefore, according to whether debiasing methods need to know the bias existing on the datasets beforehand, we group them into two types. 
(1) Known Bias-based (\text{KB-based}) methods, such as LMH~\cite{clark2019don}, LRS~\cite{guo2020loss}, and LMH-MFE~\cite{gat2020removing}. 
(2) Unknown Bias-based (\text{UB-based}) methods, such as MCE~\cite{clark2020learning}, MUTANT~\cite{gokhale2020mutant}, Unshuffling~\cite{teney2020unshuffling}, and VILLA~\cite{gan2020large}. 
Particularly, VILLA~\cite{gan2020large} utilizes the adversarial training strategy that introduces noises into representation space to improve model generalization ability. 
The strategy of injecting adversarial perturbations into representation space inspired our work. 
Therefore, we reimplement VILLA~\cite{gan2020large} on VQA-CP v2 for a fairer comparison. 
On VQA-CP v2, we utilize LXMERT and UpDn as baseline VQA models and mainly compare the proposed $\ModelName$ scheme with UB-based methods (more comparisons with state-of-the-art methods are shown in the Appendix). 
As with ablation study, we implement our experiments on VQA-CP v2 ``All'' but focus comparisons on ``Other'', which can better evaluate the ability of approaches to OOD generalization than ``Yes/No'' and ``Num.''. 

\noindentparagraph{\textbf{Results.}} 
Comparisons are reported in Table~\ref{tab:vqacpv2}. 
As shown in the table, facilitated by the proposed $\ModelName$ scheme, the baseline VQA model (LXMERT) achieves state-of-the-art OOD and ID performance on VQA-CP v2 ``Other''. 
Besides, compared with all methods utilized UpDn as the baseline VQA model, our method also gains the best OOD and ID performance on ``Other''. 
These results demonstrate the $\ModelName$ scheme is effective for a VQA model to generalize to out-of-distribution samples while preserving its ID performance. 

\subsubsection{Performance on GQA-OOD} 
\begin{table}[!tbp]
\begin{center}
\caption{Comparisons with state-of-the-arts on GQA-OOD. 
}
\vspace{-3mm}
\label{tab:gqa_ood}
\setlength{\tabcolsep}{2.8mm}{
\centering
\begin{tabular}{|l|c|c|c|c|}
\hline
\multirow{2}{*}{Method}
&\multicolumn{4}{c|}{Metrics (\%)} 
\\
\cline{2-5}
&All \textcolor{blue}{$\uparrow$}
&\text{Tail} \textcolor{blue}{$\uparrow$}
&Head \textcolor{blue}{$\uparrow$}
&$\Delta$ \textcolor{red}{$\downarrow$} \\ 
\hline
\hline
\multicolumn{5}{|l|}{\textbf{\small Plain Models}} \\
\hline 
BAN4~\cite{kim2018bilinear} &50.20 
&{47.20} &51.90 &9.90 \\ 
MMN~\cite{chen2021meta} &52.70 
&{48.00} &55.50 &15.60 \\ 
\hline
UpDn~\cite{anderson2018bottom} &46.40 
&{42.10} &49.10 &16.60 \\
\quad + LMH~\cite{clark2019don} &33.10 
&{30.80} &34.50 &12.00 \\
\quad + LM~\cite{clark2019don} &34.50 
&{32.20} &35.90 &11.50 \\ 

\quad + RUBi~\cite{cadene2019rubi} &38.80 
&{35.70} &40.80 &14.30 \\

\quad + $\ModelName$ &48.41 
&{45.99} &{49.90} &\textbf{8.51} \\
\hline
\hline

\multicolumn{5}{|l|}{\textbf{\small Transformer-based Models}} \\
\hline 
MCAN~\cite{yu2019deep} &50.80 
&{46.50} &53.40 &14.80 \\  
MANGO\footnotemark[3]~\cite{li2020closer} &\textbf{56.40} 
&51.27 &\textbf{59.55} &16.15 \\ 
\hline
LXMERT~\cite{tan2019lxmert}
&54.60 &{49.80} &\text{57.70} &15.90 \\ 
\quad + VILLA$^{*}$~\cite{gan2020large} 
&54.47 
&{49.95} &57.24 &14.59 \\
\quad + MANGO~\cite{li2020closer} 
&54.94 &{---} &--- &--- \\

\quad + $\ModelName$ 
&\text{55.59} 
&{\textbf{52.47}} &57.51 &\text{9.59} \\ 
\hline
\end{tabular}    
}
\end{center}
\vspace{-2mm}
\end{table}

\noindentparagraph{\textbf{Settings.}}
Except for LMH~\cite{clark2019don} and LM~\cite{clark2019don}, 
the other methods listed in Table~\ref{tab:gqa_ood} are UB-based and can be further grouped into plain models (models in the third row of Table~\ref{tab:gqa_ood}, \eg, BAN4~\cite{kim2018bilinear} and MMN~\cite{chen2021meta}) and transformer-based models (models in the fourth row of Table~\ref{tab:gqa_ood} like MCAN~\cite{yu2019deep} and VILLA~\cite{gan2020large}). 
We utilize LXMERT and UpDn as baseline VQA models, and compare the proposed $\ModelName$ scheme with all existing methods on GQA-OOD. 
Specifically, RUBi~\cite{cadene2019rubi}, LMH~\cite{clark2019don} and LM~\cite{clark2019don} utilize UpDn as their baseline model. 
VILLA~\cite{gan2020large} and MANGO~\cite{li2020closer} use LXMERT as the baseline VQA model. 
We also reimplement VILLA on GQA-OOD. 

\noindentparagraph{\textbf{Results.}} 
From the comparisons in Table~\ref{tab:gqa_ood}, we can observe that benefiting from the OOD generalization ability of $\ModelName$, the baseline VQA model (LXMERT) trained with our scheme outperforms all other methods on GQA-OOD on the metric \textbf{Tail}. 
Although the ID performance (Head) drops by 0.19 relative to the baseline VQA model, it is far smaller than the OOD performance improvement (2.67). 
Besides, the baseline VQA model (UpDn) trained with the $\ModelName$ scheme also outperforms these methods utilized UpDn as the baseline VQA model on all metrics, especially on \textbf{Tail}. 
These results demonstrate the ability of the proposed $\ModelName$ scheme to OOD generalization and ID preserving. 
Unfortunately, the OOD performance improvement on GQA-OOD is not as obvious as that on VQA-CP v2. 
It likely because questions of GQA-OOD are semantic compositionality. 
The syntax-complex questions weaken the representation ability of $\xv$ while $\ModelName$ also does not explicitly leverage the structured representation of question to enhance $\xv$. 

\footnotetext[3]{using UNITER~\cite{chen2020uniter} as the baseline VQA model.}

\subsection{Qualitative Examples}

\begin{figure}[!tbp]
\begin{center}
\includegraphics[width=1.0\linewidth]{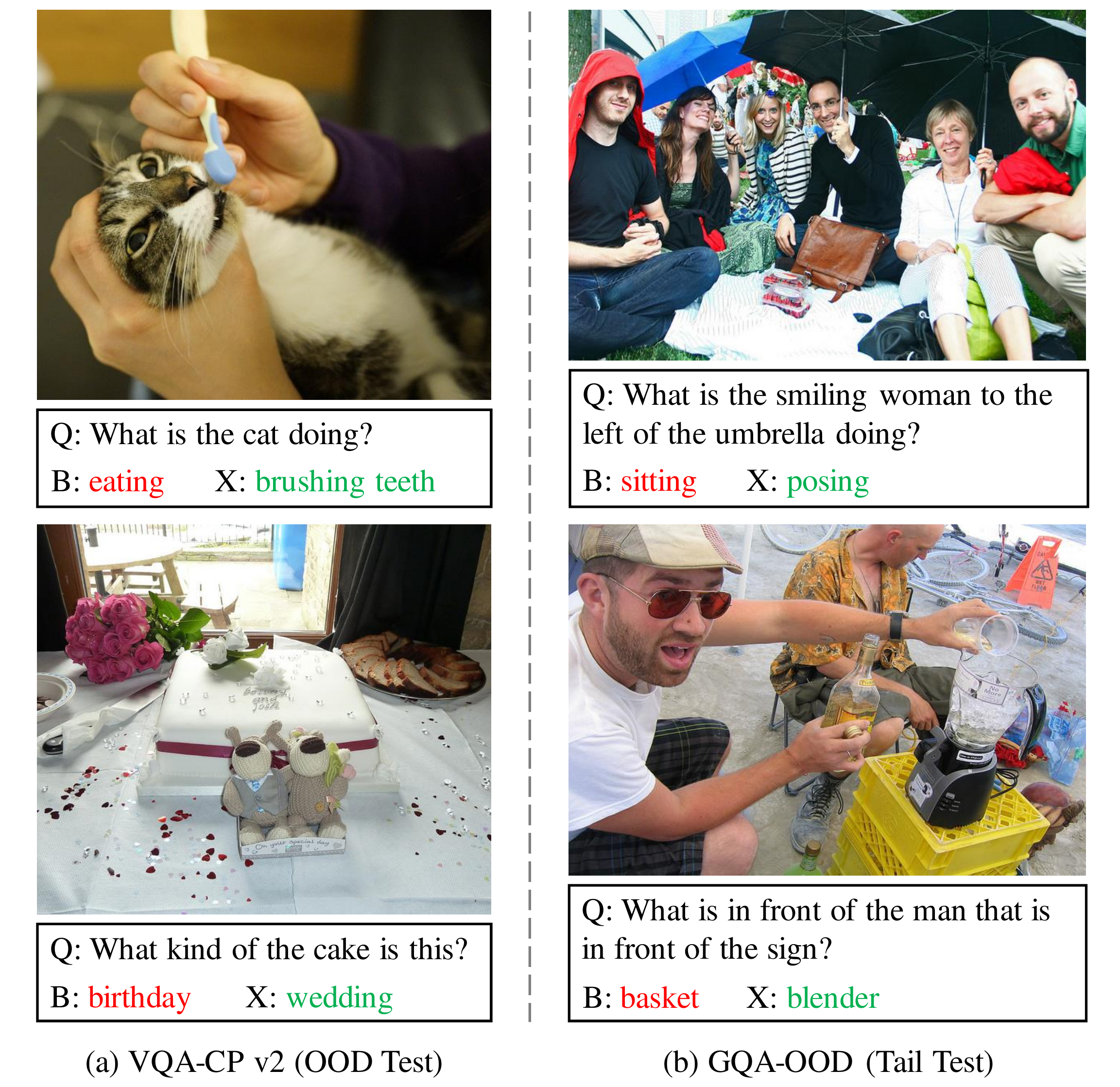}
\end{center}
\vspace{-2mm}
\caption{
Qualitative examples from the OOD test set of VQA-CP v2 and GQA-OOD respectively provided by the Baseline (B), \ie, LXMERT, and the Baseline trained with the $\ModelName$ scheme (X). 
The \textcolor{red}{wrong} and \textcolor{ForestGreen}{correct} answers are highlighted in \textcolor{red}{red} and \textcolor{ForestGreen}{green}, respectively. 
}
\label{fig:vis_prediction}
\Description{Vis.}
\vspace{-2mm}
\end{figure}

To qualitatively illustrate the effectiveness of the $\ModelName$ scheme in improving the OOD performance of baseline VQA models, we show some prediction examples from the OOD test sets of VQA-CP v2 and GQA-OOD. 
As shown in Figure~\ref{fig:vis_prediction}, the Baseline (\ie, LXMERT) is prone to utilize in-domain answers (\eg, \textit{eating} and \textit{sitting}) to answer out-of-distribution questions. 
In contrast, the Baseline after being trained with the proposed $\ModelName$ scheme can yield out-of-distribution answers, such as \textit{brushing teeth} and \textit{posing} in the first row of Figure~\ref{fig:vis_prediction}, demonstrating the ability of $\ModelName$ to OOD generalization. 
More predictions and visualization of the generated relation matrices are shown in Figure~\ref{fig:additional_vis_prediction}-\ref{fig:vis_Rmat_additional} of the Appendix. 

\section{Conclusion} 
In this paper, we proposed a Graph Generative Modeling-based scheme, $\ModelName$, to improve the OOD generalization ability of baseline VQA models while preserving the ID performance. 
Specifically, the $\ModelName$ scheme randomly executes $\RGenerator$ or $\NGenerator$ to generate a new relation matrix or new node representations to enable the baseline VQA model to generalize to out-of-distribution samples. 
Besides, to alleviate the unstable gradient issue in graph adversarial learning, we propose a gradient distribution consistency loss to constrain the derivative of data distribution with adversarial perturbations and generated distribution. 
We evaluate the effectiveness of $\ModelName$ components by extensive and ablative experiments. 
The baseline VQA model (LXMERT) trained with the proposed $\ModelName$ scheme achieves state-of-the-art OOD performance on two VQA OOD benchmarks while preserving the ID performance.

\begin{acks}
This work was supported by the NSFC Grants 61773312, 62088102. 
\end{acks}

\bibliographystyle{ACM-Reference-Format}
\bibliography{./bib/Reference}

\appendix 
\section{Appendix} 

\subsection{More Experimental Results} 

\noindent{\bf Ablation Studies.} 
The results in Table~\ref{tab:abl_comps} indicate that a single $\RGenerator$ or $\NGenerator$ can also effectively improve the OOD generalization ability of baseline VQA models in addition to the full $\ModelName$ scheme, and $\mathcal{L}_{\nabla}$ is more effective than $\mathcal{L}_D$ for alleviating the unstable gradient issue when training the baseline VQA model with the $\ModelName$ scheme. 
To further illustrate the effectiveness of $\mathcal{L}_{\nabla}$ when training the Baseline with single $\RGenerator$ or $\NGenerator$, as shown in Table~\ref{tab:abl_comps_supp} (\#6, \#7, \#8, \#9), we conduct more ablated experiments. 
Specifically, 
$\vartriangleright$ \#6 only using $\mathcal{L}_{\nabla}$ as the adversarial loss to train the baseline VQA model (LXMERT) with the $\RGenerator$. 
$\vartriangleright$ \#7 only using $\mathcal{L}_D$ as the adversarial loss to train the Baseline with the $\RGenerator$. 
$\vartriangleright$ \#8 only using $\mathcal{L}_{\nabla}$ as the adversarial loss to train the Baseline with the $\NGenerator$.  
$\vartriangleright$ \#9 only using $\mathcal{L}_D$ as the adversarial loss to train the Baseline with the $\NGenerator$. 
The results on VQA-CP v2 ``Other'' and GQA-OOD are shown in Table~\ref{tab:abl_comps_supp}, which consistently illustrate that $\mathcal{L}_{\nabla}$ is more effective than $\mathcal{L}_D$ when training the baseline VQA model with a single $\RGenerator$ or $\NGenerator$.

\begin{table}[!t]
\begin{center}
\caption{Full ablation studies on $\RGenerator$ and $\NGenerator$ with different combinations of $\mathcal{L}_{\nabla}$ and $\mathcal{L}_D$. 
}
\vspace{-4mm}
\label{tab:abl_comps_supp}
\setlength{\tabcolsep}{0.95mm}{
\renewcommand\arraystretch{1.2}
\begin{tabular}{c|cc|cc|ccc|ccc}
\hline
\multirow{2}{*}{{S/N}}
&\multicolumn{2}{c|}{{$\RGenerator$}}
&\multicolumn{2}{c|}{{$\NGenerator$}} 
&\multicolumn{3}{c|}{VQA-CP v2}
&\multicolumn{3}{c}{GQA-OOD} 
\\
\cline{2-11}
&\small{$\mathcal{L}_{\nabla}$} 
&\small{$\mathcal{L}_D$} 
&\small{$\mathcal{L}_{\nabla}$}
&\small{$\mathcal{L}_D$}  
&\small OOD\textcolor{blue}{$\uparrow$}
&\small ID\textcolor{blue}{$\uparrow$}
&\small Gap\textcolor{red}{$\downarrow$} 
&\small Tail\textcolor{blue}{$\uparrow$} 
&\small Head\textcolor{blue}{$\uparrow$} 
&$\Delta$\textcolor{red}{$\downarrow$}
\\
\hline
\hline
\rowcolor{gray!15}\small{LX.} 
&\multicolumn{4}{c|}{---} 
&59.98 &65.21 &5.23 
&49.80 &57.70 &15.90 \\ 
\hline 
\rowcolor{yellow!10}
\#6 &\checkmark & & & 
&63.37 &65.71 &2.34 
&51.36 &56.95 &10.88 
\\
\rowcolor{yellow!10}
\#7 & &\checkmark & & 
&62.67 &65.21 &2.54 
&50.80 &57.13 &12.45 
\\

\#1 &\checkmark &\checkmark & & 
&63.84 &66.41 &2.57 
&51.39 &56.93 &10.77 
\\
\hline
\rowcolor{yellow!10}
\#8 & & &\checkmark &
&62.77 &65.56 &2.79 
&51.08 &57.93 &13.41 
\\
\rowcolor{yellow!10}
\#9 & & & &\checkmark 
&62.00 &65.14 &3.14 
&50.99 &57.76 &13.28 
\\
\#2 & & &\checkmark &\checkmark 
&62.90 &65.57 &2.67
&51.27 &57.30 &11.76
\\
\hline 
\#3 &\checkmark & &\checkmark & 
&63.75 &66.03 &2.28 
&51.65 &57.13 &10.61 
\\ 
\#4 & &\checkmark & &\checkmark 
&62.73 &65.99 &3.26 
&51.27 &57.19 &11.54 
\\
\hline 
\#5 &\checkmark &\checkmark &\checkmark &\checkmark 
&64.33 &66.92 &2.59 
&52.47 &57.51 &9.59 
\\
\hline
\end{tabular}
} 
\end{center}
\vspace{-4mm}
\end{table}

\begin{figure}[!t]
\begin{center}
\includegraphics[width=0.9\linewidth]{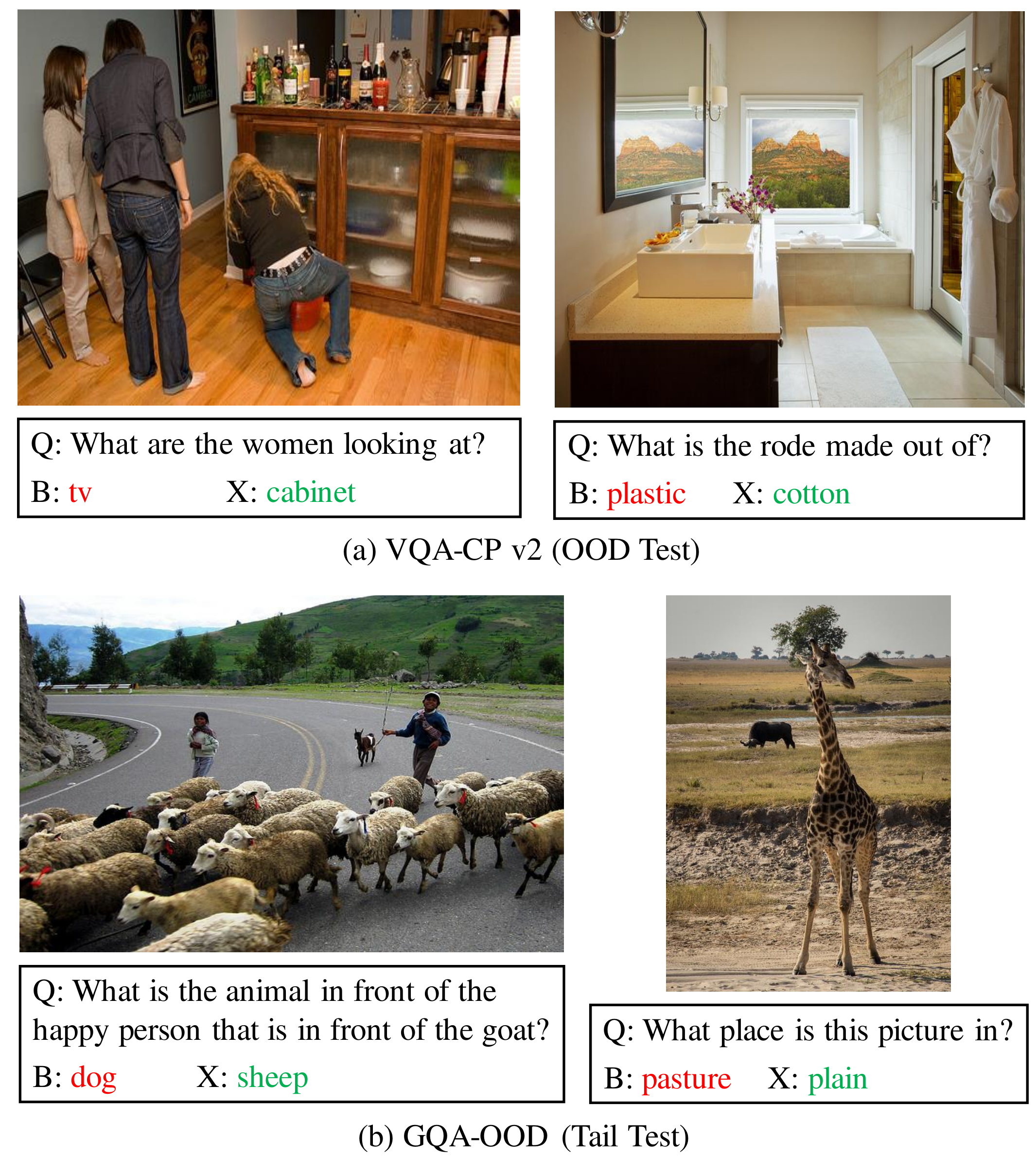}
\end{center}
\vspace{-5mm}
\caption{
Qualitative examples respectively provided by the Baseline (B), \ie, LXMERT, and the Baseline trained with the $\ModelName$ scheme (X). 
The \textcolor{red}{wrong} and \textcolor{ForestGreen}{correct} answers are highlighted in \textcolor{red}{red} and \textcolor{ForestGreen}{green}, respectively. 
} 
\label{fig:additional_vis_prediction}
\Description{Vis.}
\vspace{-4mm}
\end{figure}

\noindent{\bf Comparisons with SoTAs on VQA-CP v2.} 
Table~\ref{tab:vqacpv2_SoTA_full} shows the full comparisons with state-of-the-art debiasing methods on VQA-CP v2. 
Compared with both KB-based and UB-based methods, the baseline VQA model (LXMERT) trained with the proposed $\ModelName$ scheme can achieve state-of-the-art OOD and ID performance on  VQA-CP v2 ``Other'', further demonstrating the effectiveness of $\ModelName$ to OOD generalization.  

\noindent{\bf Additional Qualitative Examples.} 
Like Figure~\ref{fig:vis_prediction}, Figure~\ref{fig:additional_vis_prediction} shows some additional qualitative examples from the OOD test set of VQA-CP v2 and GQA-OOD provided by the Baseline (\ie, LXMERT) and the Baseline trained with the $\ModelName$ scheme. 

\begin{figure*}[!t]
\begin{center}
\includegraphics[width=1.0\linewidth]{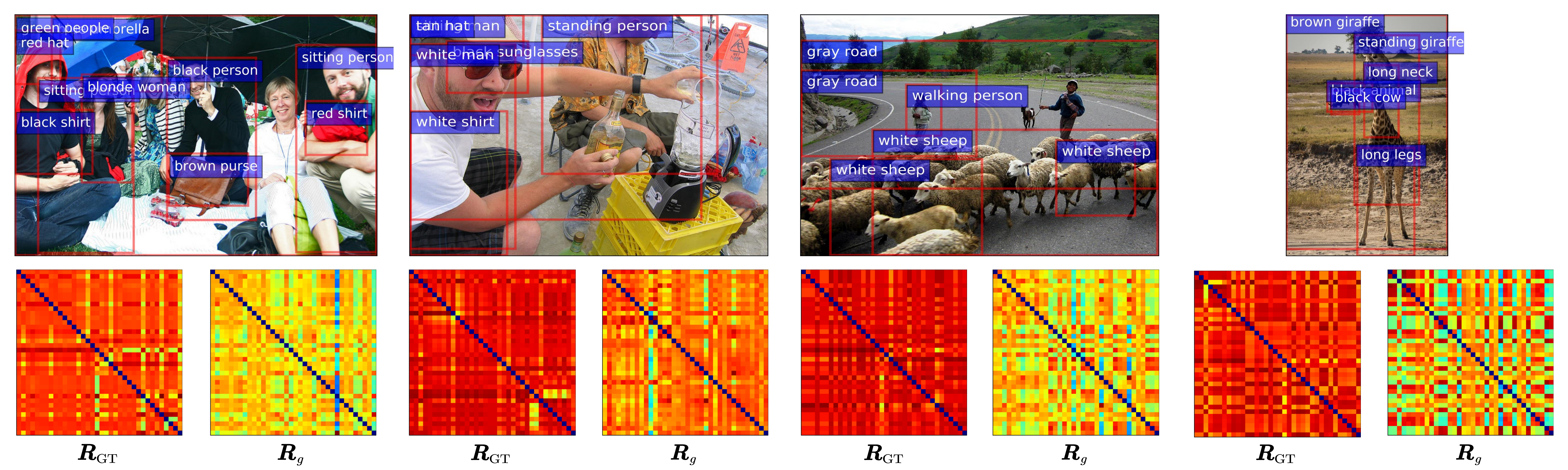}
\end{center}
\caption{
Heat maps of relation matrices. 
$\Rmat_{GT}$ is the pre-defined relation matrix (\ie, the cosine similarity between object-class embeddings and object-attribute embeddings on the image), $\Rmat_{g}$ is the generated relation matrix during $\RGenerator$. 
}
\label{fig:vis_Rmat_additional}
\Description{Vis.}
\end{figure*}

\noindent{\bf Visualization of Relation Matrix.} 
To further illustrate the process of graph generative modeling, considering the direct relation generation (\ie, $\RGenerator$), Figure~\ref{fig:vis_Rmat_additional} compares the head maps of the pre-defined relation matrix $\Rmat_{GT}$ and the generated relation matrix $\Rmat_{g}$ of images which originate from the test set of GQA-OOD in Figure~\ref{fig:vis_prediction} and Figure~\ref{fig:additional_vis_prediction}. 
From the comparison between the heat maps of the pre-fined relation matrix and the generated one, we observe that correlations between nodes becomes more discriminative and new correlations are yielded after $\RGenerator$. 

\begin{table*}[!t]
\begin{center}
\caption{
Comparisons with state-of-the-art debiasing methods on VQA-CP v2. 
$^{\dagger}$: methods using additional manually annotated data, $^{*}$: results from our reimplementation using the official released code, ---: not reported. 
}
\label{tab:vqacpv2_SoTA_full}
\centering 
\renewcommand\arraystretch{1.1}
\begin{tabular}{|l|cccc|cccc|cccc|}
\hline 
\multirow{2}{*}{Method} 
&\multicolumn{4}{c|}{VQA-CP v2 (OOD Test) \textcolor{blue}{$\uparrow$}} 
&\multicolumn{4}{c|}{VQA-CP v2 (ID Test) \textcolor{blue}{$\uparrow$}} 
&\multicolumn{4}{c|}{Gap \textcolor{red}{$\downarrow$}} \\ 
\cline{2-13}
&All &Yes/No &Num. &Other 
&All &Yes/No &Num. &Other 
&All &Yes/No &Num. &Other \\ 
\hline 
\hline
\multicolumn{13}{|l|}{\small \textbf{KB-based Methods}} \\
\hline 
RankVQA~\cite{qiao2020rankvqa} 
&43.05 &42.53 &13.19 &\cellcolor{cyan!10}{51.32} 
&--- &--- &--- &\cellcolor{cyan!10}{---} 
&--- &--- &--- &\cellcolor{cyan!10}{---} \\
\hline
LM~\cite{clark2019don} 
&48.78 &72.78 &14.61 &\cellcolor{cyan!10}{45.58} 
&--- &--- &--- &\cellcolor{cyan!10}{---} 
&--- &--- &--- &\cellcolor{cyan!10}{---} \\
LMH~\cite{clark2019don} 
&52.01 &72.58 &31.12 &\cellcolor{cyan!10}{46.97} 
&--- &--- &--- &\cellcolor{cyan!10}{---} 
&--- &--- &--- &\cellcolor{cyan!10}{---} \\
\quad + LRS~\cite{guo2020loss} 
&53.26 &72.82 &48.00 &\cellcolor{cyan!10}{44.46}
&--- &--- &--- &\cellcolor{cyan!10}{---} 
&--- &--- &--- &\cellcolor{cyan!10}{---}  \\
\quad + CCB+VQ-CSS~\cite{yang2020learning} 
&59.12 &89.12 &51.04 &\cellcolor{cyan!10}{45.62} 
&--- &--- &--- &\cellcolor{cyan!10}{---} 
&--- &--- &--- &\cellcolor{cyan!10}{---} \\
\quad + MFE~\cite{gat2020removing} 
&54.55 &74.03 &49.16 &\cellcolor{cyan!10}{45.82} 
&--- &--- &--- &\cellcolor{cyan!10}{---} 
&--- &--- &--- &\cellcolor{cyan!10}{---} \\ 
\quad + CSS+CL~\cite{liang2020learning} 
&59.18 &86.99 &49.89 &\cellcolor{cyan!10}{47.16} 
&--- &--- &--- &\cellcolor{cyan!10}{---} 
&--- &--- &---&\cellcolor{cyan!10}{---} \\
\quad + CSS+GS~\cite{teney2020learning,liang2020learning}
&57.37 &79.71 &50.85 &\cellcolor{cyan!10}{47.45} 
&--- &--- &--- &\cellcolor{cyan!10}{---} 
&--- &--- &---&\cellcolor{cyan!10}{---} \\
\quad + CSS~\cite{chen2020counterfactual} 
&58.95 &84.37 &49.42 &\cellcolor{cyan!10}{48.21} 
&--- &--- &--- &\cellcolor{cyan!10}{---} 
& &--- &---&\cellcolor{cyan!10}{---} \\ 
\quad + CCB~\cite{yang2020learning} 
&57.99 &86.41 &45.63 &\cellcolor{cyan!10}{48.76} 
&--- &--- &--- &\cellcolor{cyan!10}{---} 
&--- &--- &---&\cellcolor{cyan!10}{---} \\
\quad + MCE (LXMERT)~\cite{clark2020learning} 
&\text{70.32} &--- &--- &\cellcolor{cyan!10}{---} 
&70.78 &--- &--- &\cellcolor{cyan!10}{---} 
&0.46 &--- &--- &\cellcolor{cyan!10}{---} \\
\hline 
\hline
\multicolumn{13}{|l|}{\textbf{\small UB-based Methods}} \\
\hline
GVQA~\cite{agrawal2018don} 
&31.30 &57.99 &13.68 &\cellcolor{cyan!10}{22.14}
&--- &--- &--- &\cellcolor{cyan!10}{}{---} 
&--- &--- &--- &\cellcolor{cyan!10}{---} \\ 
Actively Seeking~\cite{teney2019actively} 
&46.00 &58.24 &29.49 &\cellcolor{cyan!10}{44.33} 
&--- &--- &--- &\cellcolor{cyan!10}{---} 
&--- &--- &--- &\cellcolor{cyan!10}{---} \\
Unshuffling+CF+GS~\cite{teney2020learning}
&46.80 &64.50 &15.30 &\cellcolor{cyan!10}{45.90}
&62.40 &77.80 &43.80 &\cellcolor{cyan!10}{53.60}
&15.60 &13.30 &28.50 &\cellcolor{cyan!10}{7.70} \\ 
Unshuffling+CF~\cite{teney2020learning} 
&40.60 &61.30 &15.60 &\cellcolor{cyan!10}{46.00} 
&63.30 &79.40 &45.50 &\cellcolor{cyan!10}{53.70} 
&22.70 &18.10 &29.90 &\cellcolor{cyan!10}{7.70} \\
VGQE~\cite{kv2020reducing} 
&50.11 &66.35 &27.08 &\cellcolor{cyan!10}{46.77} 
&--- &--- &--- &\cellcolor{cyan!10}{---} 
&--- &--- &--- &\cellcolor{cyan!10}{---} \\ 
Unshuffling~\cite{teney2020unshuffling} 
&42.39 &47.72 &14.43 &\cellcolor{cyan!10}{47.24} 
&--- &--- &--- &\cellcolor{cyan!10}{---} 
&--- &--- &--- &\cellcolor{cyan!10}{---} \\ 
\hline
UpDn~\cite{anderson2018bottom} 
&38.82  &42.98  &12.18  &\cellcolor{cyan!10}{43.95} 
&64.73  &79.45  &49.59  &\cellcolor{cyan!10}{55.66}  
&25.91  &36.47 &37.41 &\cellcolor{cyan!10}{11.71} \\

\quad + Top answer masked~\cite{teney2020value} 
&40.61  &82.44  &27.63  &\cellcolor{cyan!10}{22.26} 
&30.90  &44.12  &5.00  &\cellcolor{cyan!10}{20.85} 
&9.71 &38.32 &22.63 &\cellcolor{cyan!10}{{\textbf{1.41}}} \\ 
\quad + AdvReg~\cite{grand2019adversarial} 
&36.33  &59.33  &14.01  &\cellcolor{cyan!10}{30.41} 
&50.63  &67.39  &38.81  &\cellcolor{cyan!10}{38.37} 
&14.30 &8.06 &24.80 &\cellcolor{cyan!10}{7.96} \\ 
\quad + GRL~\cite{grand2019adversarial}
&42.33 &59.74 &14.78 &\cellcolor{cyan!10}{40.76} 
&56.90 &69.23 &42.50 &\cellcolor{cyan!10}{49.36} 
&14.57 &9.49 &27.72 &\cellcolor{cyan!10}{8.60} \\
\quad + RUBi~\cite{cadene2019rubi} 
&47.11 &68.65 &20.28 &\cellcolor{cyan!10}{43.18} 
&--- &--- &--- &\cellcolor{cyan!10}{---} 
&--- &--- &--- &\cellcolor{cyan!10}{---} \\
\quad + RandImg ($\lambda$=12)~\cite{teney2020value} 
&55.37  &83.89  &41.60  &\cellcolor{cyan!10}{44.20}  
&54.24  &64.22  &34.40  &\cellcolor{cyan!10}{50.46}
&\text{1.13} &19.67 &7.20 &\cellcolor{cyan!10}{6.26} \\
\quad + CF-VQA (Sum)~\cite{niu2020counterfactual}
&53.69 &91.25 &12.80 &\cellcolor{cyan!10}{45.23} 
&--- &--- &--- &\cellcolor{cyan!10}{---}  
&--- &--- &--- &\cellcolor{cyan!10}{---} \\
\quad + CF-VQA (Har.)~\cite{niu2020counterfactual}
&49.94 &74.82 &18.93 &\cellcolor{cyan!10}{45.42} 
&--- &--- &--- &\cellcolor{cyan!10}{---}  
&--- &--- &--- &\cellcolor{cyan!10}{---} \\
\quad + DLR~\cite{jing2020overcoming} 
&48.87 &70.99 &18.72 &\cellcolor{cyan!10}{45.57}  
&--- &--- &--- &\cellcolor{cyan!10}{---} 
&--- &--- &--- &\cellcolor{cyan!10}{---}  \\
\quad + RandImg ($\lambda$=5)~\cite{teney2020value}
&51.15  &75.06  &24.30  &\cellcolor{cyan!10}{45.99}  
&59.28  &70.66  &43.06  &\cellcolor{cyan!10}{53.40} 
&{8.13} &4.40 &18.76 &\cellcolor{cyan!10}{7.41} \\ 
\quad + HINT(HAT)$^{\dagger}$~\cite{selvaraju2019taking} 
&47.70 &70.04 &10.68 &\cellcolor{cyan!10}{46.31} 
&--- &--- &--- &\cellcolor{cyan!10}{---} 
&--- &--- &--- &\cellcolor{cyan!10}{---} \\
\quad + SCR (HAT)$^{\dagger}$~\cite{wu2019self} 
&49.17 &71.55 &10.72 &\cellcolor{cyan!10}{47.49} 
&--- &--- &--- &\cellcolor{cyan!10}{---} 
&--- &--- &--- &\cellcolor{cyan!10}{---} \\
\quad + SCR (VQA-X)$^{\dagger}$~\cite{wu2019self} 
&49.45 &72.36 &10.93 &\cellcolor{cyan!10}{48.02} 
&--- &--- &--- &\cellcolor{cyan!10}{---} 
&--- &--- &--- &\cellcolor{cyan!10}{---} \\
\quad + SL~\cite{zhu2020overcoming} 
&57.59 &86.53 &29.87 &\cellcolor{cyan!10}{50.03}  
&--- &--- &--- &\cellcolor{cyan!10}{---} 
&--- &--- &--- &\cellcolor{cyan!10}{---} \\
\quad + MUTANT$^{\dagger}$~\cite{gokhale2020mutant}
&61.72 &88.90 &49.68 &\cellcolor{cyan!10}{50.78} 
&--- &--- &--- &\cellcolor{cyan!10}{---} 
&--- &--- &--- &\cellcolor{cyan!10}{---} \\

\quad + $\ModelName$ 
&45.71  &43.48  &27.65  &\cellcolor{cyan!10}{{52.34}}   
&67.16  &83.74  &48.26  &\cellcolor{cyan!10}{{56.91}} 
&21.45  &40.26  &20.61  &\cellcolor{cyan!10}{\text{4.57}} \\ 

\hline
MANGO~\cite{li2020closer} 
&52.76 &--- &--- &\cellcolor{cyan!10}{---}
&--- &--- &--- &\cellcolor{cyan!10}{---} 
&--- &--- &--- &\cellcolor{cyan!10}{---} \\
\hline
LXMERT$^{*}$~\cite{tan2019lxmert} 
&63.90 &80.45 &46.58 &\cellcolor{cyan!10}{{59.98}} 
&75.57 &91.39 &59.83 &\cellcolor{cyan!10}{65.21}
&11.67 &10.94 &13.25 &\cellcolor{cyan!10}{5.23} \\
\quad + MUTANT$^{\dagger}$~\cite{gokhale2020mutant}
&{\text{69.52}} &93.15 &67.17 &\cellcolor{cyan!10}{57.78} 
&--- &--- &--- &\cellcolor{cyan!10}{---} 
&--- &--- &--- &\cellcolor{cyan!10}{---} \\
\quad + VILLA$^*$~\cite{gan2020large} 
&48.06 &42.66 &18.97 &\cellcolor{cyan!10}{{58.89}} 
&{75.85} &90.40 &60.02 &\cellcolor{cyan!10}{{66.66}} 
&27.79 &47.97 &41.05 &\cellcolor{cyan!10}{7.77} \\

\quad + MCE+Adv~\cite{clark2020learning} 
&66.08 &--- &--- &\cellcolor{cyan!10}{---}
&{\text{77.17}} &--- &--- &\cellcolor{cyan!10}{---}
&11.09 &--- &--- &\cellcolor{cyan!10}{---} \\
\quad + MCE~\cite{clark2020learning} 
&{68.44} &--- &--- &\cellcolor{cyan!10}{---}
&74.03 &--- &--- &\cellcolor{cyan!10}{---}
&{5.59} &--- &--- &\cellcolor{cyan!10}{---} \\

\quad + $\ModelName$ 
&66.95 &78.91 &53.63 &\cellcolor{cyan!10}{{\textbf{64.33}}} 
&75.88 &90.18 &60.06 &\cellcolor{cyan!10}{{\textbf{66.92}}} 
&8.93 &11.27 &6.43 &\cellcolor{cyan!10}{{{2.59}}} \\
\hline
\end{tabular}
\end{center}
\end{table*}

\end{document}